\documentclass{article}

\usepackage{microtype}
\usepackage{graphicx}
\usepackage{subfigure}
\usepackage{booktabs} 

\usepackage{hyperref}

\usepackage[accepted]{icml2024}

\usepackage{amsmath}
\usepackage{amssymb}
\usepackage{mathtools}
\usepackage{amsthm}
\usepackage{zymacros}

\usepackage[capitalize,noabbrev]{cleveref}

\theoremstyle{plain}
\newtheorem{theorem}{Theorem}[section]
\newtheorem{proposition}[theorem]{Proposition}
\newtheorem{lemma}[theorem]{Lemma}
\newtheorem{corollary}[theorem]{Corollary}
\theoremstyle{definition}
\newtheorem{definition}[theorem]{Definition}

\theoremstyle{remark}
\newtheorem{remark}[theorem]{Remark}

\usepackage[textsize=tiny]{todonotes}

\icmltitlerunning{Deeper or Wider: A Perspective from Optimal Generalization Error with Sobolev Loss}

\begin{document}

\twocolumn[
\icmltitle{Deeper or Wider: A Perspective from \\ Optimal Generalization Error with Sobolev Loss}




\begin{icmlauthorlist}
\icmlauthor{Yahong Yang}{psu}
\icmlauthor{Juncai He}{kaust}
\end{icmlauthorlist}

\icmlaffiliation{psu}{Department of Mathematics,
	The Pennsylvania State University, University Park, State College, PA 16802, USA}
\icmlaffiliation{kaust}{Computer, Electrical and Mathematical Science and Engineering Division,
The King Abdullah University of Science and Technology,
Thuwal 23955, Saudi Arabia}

\icmlcorrespondingauthor{Yahong Yang}{yxy5498@psu.edu}
\icmlcorrespondingauthor{Juncai He}{juncai.he@kaust.edu.sa}

\icmlkeywords{Machine Learning, ICML}

\vskip 0.3in
]



\printAffiliationsAndNotice{}  

\begin{abstract}
 Constructing the architecture of a neural network is a challenging pursuit for the machine learning community, and the dilemma of whether to go deeper or wider remains a persistent question. This paper explores a comparison between deeper neural networks (DeNNs) with a flexible number of layers and wider neural networks (WeNNs) with limited hidden layers, focusing on their optimal generalization error in Sobolev losses. Analytical investigations reveal that the architecture of a neural network can be significantly influenced by various factors, including the number of sample points, parameters within the neural networks, and the regularity of the loss function. Specifically, a higher number of parameters tends to favor WeNNs, while an increased number of sample points and greater regularity in the loss function lean towards the adoption of DeNNs. We ultimately apply this theory to address partial differential equations using deep Ritz and physics-informed neural network (PINN) methods, guiding the design of neural networks.

\end{abstract}

\section{Introduction}
\label{intro}

Recently, Sobolev training \cite{czarnecki2017sobolev, son2021sobolev, vlassis2021sobolev} for neural networks (NNs) with the rectified linear unit (ReLU) activation function has improved model prediction and generalization capabilities by incorporating derivatives of the target output with respect to the model input.  This training approach has proven impactful in various areas, including the solution of partial differential equations \cite{Lagaris1998, weinan2017deep, raissi2019physics, de2022error}, operator learning \cite{lu2021learning}, network compression \cite{sau2016deep}, distillation \cite{hinton2015distilling, rusu2015policy}, regularization \cite{czarnecki2017sobolev}, and dynamic programming \cite{finlay2018lipschitz, werbos1992approximate}, among others.

In Sobolev training, two architectural parameters of fully connected neural networks have been considered: $L$, the number of hidden layers, and $N$, the number of neurons in each layer. In this paper, we consider the following two different NNs with a fixed number of parameters $W = \mathcal O(N^2L)$:

\begin{itemize}
    \item Wider neural networks (WeNNs) with a limited number of hidden layers but flexible width, namely, $L = \mathcal O(1)$ or ${\mathcal O}\left(\log(N)\right)$;
\item Deeper neural networks (DeNNs) with a flexible number of layers but limited width: $N = \mathcal O(1)$ or ${\mathcal O}\left(\log(L)\right)$.
\end{itemize}

In the literature, WeNNs, as evidenced by recent works such as \cite{schmidt2020nonparametric, suzuki2018adaptivity, ma2022barron, barron1993universal, mhaskar1996neural, ma2022uniform, siegel2022sharp, montanelli2019new}, demonstrate efficacy in tasks that require shallower architectures. In contrast, DeNNs, supported by recent research \cite{yarotsky2017error, yarotsky2020phase, siegel2022optimal, yang2023nearly, yang2023nearlys, he2023optimal, yang2023optimal}, exhibit extensive capacity for intricate representations and excel in handling complex computations.

These two categories of neural networks have distinct advantages and disadvantages. WeNNs boast a simple structure that facilitates easy training and reduces the risk of overfitting. However, this simplicity comes at a cost, as the limited richness within this space constrains their approximation capability. Consequently, achieving a satisfactory approximation necessitates an increased number of parameters in the network structure.
On the other hand, DeNNs offer a high-complexity space, enabling accurate approximations with fewer parameters. Works such as \cite{yarotsky2017error, siegel2022optimal, yang2023nearly, yang2023nearlys, he2023optimal, yang2023optimal} demonstrate that DeNNs can achieve significantly better approximations than shallow neural networks and traditional methods. This improved approximation rate is referred to as super-convergence. However, these studies do not explicitly compare the generalization error with shallow neural networks.
Nevertheless, the heightened complexity of DeNNs introduces challenges in the learning process. Training such networks is not a straightforward task and demands a more extensive set of sample points to effectively capture the intricate relationships within the data.

In this paper, our objective is to conduct a comprehensive analysis of the optimal generalization errors between these two types of NNs in Sobolev training setup. While existing literature, such as \cite{schmidt2020nonparametric, suzuki2018adaptivity}, has extensively explored WeNNs, our focus is on the generalization error of DeNNs, breaking it down into two integral components: the approximation error and the sampling error.
Despite previous studies, including \cite{duan2021convergence, jakubovitz2019generalization, advani2020high, berner2020analysis}, analyzing the generalization error of DeNNs, these works impose a constraint requiring the parameters in DeNNs to be uniformly bounded. This constraint limits the complexity of DeNNs and compromises the approximation advantages inherent in DeNNs. Notably, without such a bounded restriction, DeNNs can achieve a superior approximation rate compared to traditional NNs \cite{yarotsky2017error,  siegel2022optimal, yang2023nearly, yang2023nearlys, he2023optimal, yang2023optimal}. Furthermore, our result is different from \cite{yang2023nearly, yang2023nearlys}, which is not nearly optimal in those papers. Moreover, in this paper, we will extend the generalization error analysis to Sobolev training to further understand the capabilities of DeNNs in a more general setting.

\begin{figure*}[h!]
    \centering
    \includegraphics[scale=0.77]{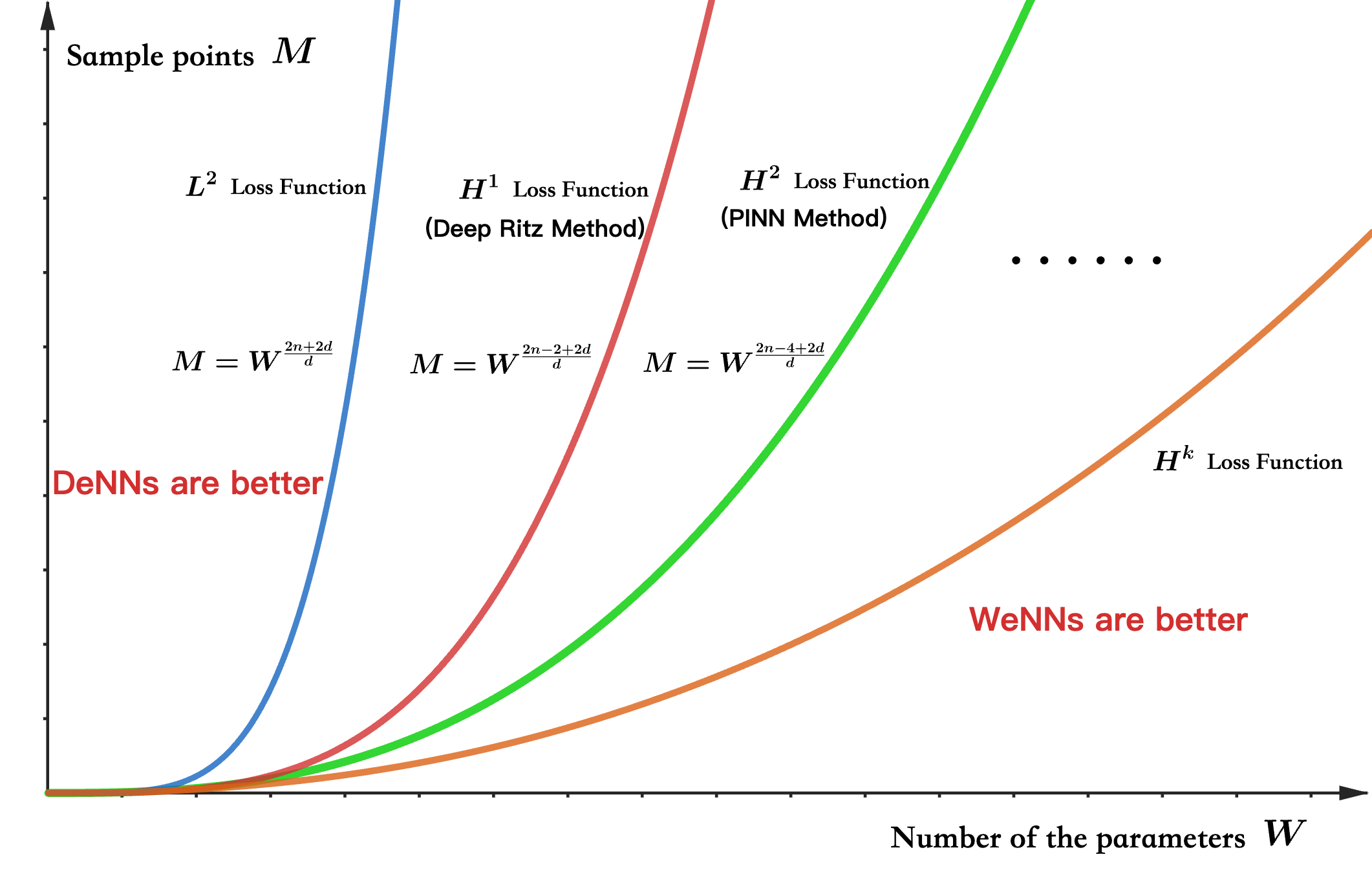}
    \caption{Generalization error with respect to the number of parameters, $W$, and sample points, $M$, for $L^2$ loss functions, $H^1$ loss functions, $H^2$ loss functions, and $H^k$ loss functions with $k\ge 3$ ($W=\fO(N^2L)$, where $N$ is the width of NNs and $L$ is the depth of NNs).}
    \label{deep_shallowall}
\end{figure*}

We investigate the approximation of the target function $f(\vx)$ defined on $[0,1]^d$ with $\|f\|_{W^{n,\infty}\left([0,1]^d\right)}\le 1$ using a finite set of sample data $\{(\vx_i,f(\vx_i))\}_{i=1}^M$. We employ DeNNs with a width of $N$ and a depth of $L$, leading to a total number of parameters denoted as $W:=\fO(N^2L)$. 
The main results of this paper are summarized in Table~\ref{table1} and Figure~\ref{deep_shallowall}. Table~\ref{table1} lists the approximation and sampling errors for both WeNNs and DeNNs with different loss functions.
 Fig.~\ref{deep_shallowall} summarizes the genearlization error with different soloblev loss functions with respect to the number of parameters $W$ and the number of sample points $M$. For instance, the blue curve represents the generalization error with $L^2$ loss function. When a pair $(W,M)$ is located to the left of this curve, we demonstrate that the DeNNs performs better. Conversely, when the pair $(W,M)$ is situated to the right of this curve, WeNNs are more effective. Similar curves are drawn for $H^k$ loss functions, revealing that WeNNs perform better  as the loss function requires higher regularity.
Here, the $H^1$  loss is equivalent to the deep Ritz method, and the $H^2$ loss represents the PINN method.

\begin{table*}
\centering
\begin{tabular}{|c|c|c|c|c|}
\hline & \multicolumn{2}{|c|}{\text { WeNNs }} & \multicolumn{2}{|c|}{\text { DeNNs }} \\
\hline & \text { Approximation } & \text { Sampling  (Optimal)} & \text { Approximation (Optimal)} & \text { Sampling (Optimal)} \\
\hline $L^2$-loss & $\fO(W^{-\frac{n}{d}})$  & $\fO(W/M)$&  $\fO(W^{-\frac{2n}{d}})$  & $\fO(W^2/M)$ \\
\hline $H^1$-loss & $\fO(W^{-\frac{n-1}{d}})$ &  $\fO(W/M)$  &  $\fO(W^{-\frac{2n-2}{d}})$  & $\fO(W^2/M)$ \\
\hline $H^2$-loss & $\fO(W^{-\frac{n-2}{d}})$ & $\fO(W/M)$ &  $\fO(W^{-\frac{2n-4}{d}})$  & $\fO(W^2/M)$ \\
\hline
\end{tabular}\caption{The approximation and sampling errors for both WeNNs and DeNNs ($W$ represents the number of parameters, defined as $W:=\fO(N^2L)$ with a width of $N$ and a depth of $L$, while $M$ denotes the number of sample points. All values in the table are up to a logarithmic factor. The results for both WeNNs and DeNNs are near optimality in terms of the sample number \cite{schmidt2020nonparametric, suzuki2018adaptivity}. Additionally, the results for DeNNs approach near optimality in terms of the parameter count $W$ \cite{lu2021deep, siegel2022optimal, yang2023nearly, yang2023nearlys}.}\label{table1}
\end{table*} 

Based on Fig.~\ref{deep_shallowall}, valuable insights for selecting neural networks can be gained. If the number of parameters in our neural networks is fixed, the choice between WeNNs and DeNNs depends on the availability of sample points. Opting for WeNNs is advisable when the number of sample points is limited, whereas choosing DeNNs becomes preferable when a substantial number of sample points are available.
Conversely, when the number of sample points is fixed, the neural network architecture will depend on the number of parameters. DeNNs are preferable if one seeks fewer parameters, while WeNNs are more suitable when a larger number of parameters is acceptable.

Moreover, it is evident that with an increase in the order of derivatives in the loss functions, the effective region for DeNNs expands. In cases where the pair $(W,M)$ falls within transitional areas between the two curves, enhancing the depth of the neural network is advisable to improve generalization accuracy.

We highlight the contributions of the present paper as follows:

$\bullet$ We establish the optimal generalization error by dissecting it into two components—approximation error and sampling error—for DeNNs. Furthermore, we extend this result to Sobolev training, specifically involving loss functions defined by $H^1$ and $H^2$ norms. Our result differs from existing works \citep{schmidt2020nonparametric, suzuki2018adaptivity} which only consider the generalization error of WNNs. \citep{duan2021convergence, yang2023nearly, yang2023nearlys} obtain the generalization error using Rademacher complexity, which is suboptimal. \citep{shi2024nonlinear,zhang2023classification,feng2021generalization} do not analyze the generalization error for Sobolev training.

$\bullet$ We compare the optimal generalization error of neural networks \cite{schmidt2020nonparametric, suzuki2018adaptivity} with our findings and conclude that DeNNs perform better than WeNNs when the sample points are abundant but the number of parameters is limited. Conversely, WeNNs are superior to DeNNs if the number of parameters is large but the number of sample points is limited. Furthermore, as the required order of the derivative in the loss function increases, WeNNs may transition to DeNNs, influencing the performance of NNs in generalization error.

$\bullet$ We utilize our findings to analyze the generalization error of deep Ritz and PINN methods when employing DeNNs with a flexible number of layers. Our approach surpasses existing work, as prior studies have solely focused on the approximation and sample errors for neural networks with a limited number of hidden layers.

\subsection{Organization of the paper} 
The remainder of the paper unfolds as follows: Section~\ref{preliminaries} provides a compilation of useful notations and definitions related to Sobolev spaces. Subsequently, in Section~\ref{losshk}, we delve into establishing the optimal generalization error of DeNNs with respect to the \(H^k\) norms for $k=0,1,2$. In Section~\ref{apply}, we apply our results to the deep Ritz method and PINN methods.

\section{Preliminaries}\label{preliminaries}
		\subsection{Neural networks}
		Let us summarize all basic notations used in the DeNNs as follows:
		
		\textbf{1}. Matrices are denoted by bold uppercase letters. For example, $\vA\in\sR^{m\times n}$ is a real matrix of size $m\times n$ and $\vA^\T$ denotes the transpose of $\vA$. Vectors are denoted by bold lowercase letters. For example, $\vv\in\sR^n$ is a column vector of size $n$. 
		
		\textbf{2}. For a $d$-dimensional multi-index $\valpha=[\alpha_1,\alpha_2,\cdots\alpha_d]\in\sN^d$, we denote several related notations as follows: 
  $(a)~ |\boldsymbol{\alpha}|=\left|\alpha_1\right|+\left|\alpha_2\right|+\cdots+\left|\alpha_d\right|$; $(b)~\boldsymbol{x}^\alpha=x_1^{\alpha_1} x_2^{\alpha_2} \cdots x_d^{\alpha_d},~ \boldsymbol{x}=\left[x_1, x_2, \cdots, x_d\right]^\T$; $ (c)~\boldsymbol{\alpha} !=\alpha_{1} ! \alpha_{2} ! \cdots \alpha_{d} !.$
		
		
\textbf{3}. Assume $\vn\in\sN_+^m$, and $f$ and $g$ are functions defined on $\sN_+^m$, then $f(\vn)=\fO(g(\vn))$ means that there exists positive $C$ independent of $\vn,f,g$ such that $f(\vn)\le Cg(\vn)$ when all entries of $\vn$ go to $+\infty$.
		
\textbf{4}. Define $\sigma_1(x):=\sigma(x)=\max\{0,x\}$ and $\sigma_2:=\sigma^2(x)$. We call the neural networks with activation function $\sigma_t$ with $t\le i$ as $\sigma_i$ neural networks ($\sigma_i$-NNs), $i=1,2$. With the abuse of notations, we define $\sigma_i:\sR^d\to\sR^d$ as $\sigma(\vx)=\left[\sigma(x_1), \cdots, \sigma(x_d)\right]^\T \in\sR^d$. for any $\vx=\left[x_1, \cdots, x_d\right]^\T$.
		
\textbf{5}. Define $L,N\in\sN_+$, $N_0=d$ and $N_{L+1}=1$, $N_i\in\sN_+$ for $i=1,2,\ldots,L$, then a $\sigma_i$-NN $\phi$ with the width $N$ and depth $L$ can be described as follows:\[\boldsymbol{x}=\tilde{\boldsymbol{h}}_0 \stackrel{W_1, b_1}{\longrightarrow} \boldsymbol{h}_1 \stackrel{\sigma_i}{\longrightarrow} \tilde{\boldsymbol{h}}_1 \ldots \stackrel{\sigma_i}{\longrightarrow} \tilde{\boldsymbol{h}}_L \stackrel{W_{L+1}, b_{L+1}}{\longrightarrow} \phi(\boldsymbol{x}),\] where $\vW_i\in\sR^{N_i\times N_{i-1}}$ and $\vb_i\in\sR^{N_i}$ are the weight matrix and the bias vector in the $i$-th linear transform in $\phi$, respectively, i.e., $\boldsymbol{h}_i:=\boldsymbol{W}_i \tilde{\boldsymbol{h}}_{i-1}+\boldsymbol{b}_i, ~\text { for } i=1, \ldots, L+1$ and $\tilde{\boldsymbol{h}}_i=\sigma_i\left(\boldsymbol{h}_i\right),\text{ for }i=1, \ldots, L.$ In this paper, an DeNN with the width $N$ and depth $L$, means (a) The maximum width of this DeNN for all hidden layers is less than or equal to $N$. (b) The number of hidden layers of this DeNN is less than or equal to $L$.

\subsection{Sobolev spaces and covering number}
		
Denote $\Omega$ as $[0,1]^d$, $D$ as the weak derivative of a single variable function, $D^{\valpha}=D^{\alpha_1}_1D^{\alpha_2}_2\ldots D^{\alpha_d}_d$ as the partial derivative of a multivariable function, where $\valpha=[\alpha_{1},\alpha_{2},\ldots,\alpha_d]^\T$ and $D_i$ is the derivative in the $i$-th variable.
		
		\begin{definition}[Sobolev Spaces \citep{evans2022partial}]
			Let $n\in\sN$ and $1\le p\le \infty$. Then we define Sobolev spaces\begin{align}W^{n, p}(\Omega):=\left\{f \in L^p(\Omega): D^{\valpha} f \in L^p(\Omega) \text { for all }|\boldsymbol{\alpha}| \leq n\right\}\notag\end{align} with \(\|f\|_{W^{n, p}(\Omega)}:=\left(\sum_{0 \leq|\alpha| \leq n}\left\|D^{\valpha} f\right\|_{L^p(\Omega)}^p\right)^{1 / p},\) if $p<\infty$, and $\|f\|_{W^{n, \infty}(\Omega)}:=\max_{0 \leq|\alpha| \leq n}\left\|D^{\valpha} f\right\|_{L^\infty(\Omega)}$.
			Furthermore, for $\vf=(f_1,f_2,\ldots,f_d)$, $\vf\in W^{1,\infty}(\Omega,\sR^d)$ if and only if $ f_i\in W^{1,\infty}(\Omega)$ for each $i=1,2,\ldots,d$ and \(\|\vf\|_{W^{1,\infty}(\Omega,\sR^d)}:=\max_{i=1,\ldots,d}\{\|f_i\|_{W^{1,\infty}(\Omega)}\}.\) When $p=2$, denote $W^{n,2}(\Omega)$ as $H^n(\Omega)$ for $n\in\sN_+$.
		\end{definition}

   \begin{definition}[covering number \cite{anthony1999neural}]
				Let $(V,\|\cdot\|)$ be a normed space, and $\Theta\in V$. $\{V_1,V_2,\ldots,V_n\}$ is an $\varepsilon$-covering of $\Theta$ if $\Theta\subset \cup_{i=1}^nB_{\varepsilon,\|\cdot\|}(V_i)$. The covering number $\fN(\varepsilon,\Theta,\|\cdot\|)$ is defined as \(\fN(\varepsilon,\Theta,\|\cdot\|):=\min \{n: \exists \varepsilon \text {-covering over } \Theta \text { of size } n\} \text {. }\)
			\end{definition}
			
			\begin{definition}[Uniform covering number \cite{anthony1999neural}]\label{uniform}
				{Suppose the $\fF$ is a class of functions from $\fX$ to $\sR$.} Given $n$ samples $\vZ_n=(z_1,\ldots,z_n)\in\fX^n$, define \[\fF|_{\vZ_n}=\{(u(z_1),\ldots,u(z_n)):u\in\fF\}.\]The uniform covering number $\fN(\varepsilon,\fF,n)$ is defined as \[\fN(\varepsilon,\fF,n)=\max_{\vZ_n\in\fX^n}\fN\left(\varepsilon, \fF|_{\vZ_n},\|\cdot\|_{\infty}\right),\]where $\fN\left(\varepsilon, \fF|_{\vZ_n},\|\cdot\|_{\infty}\right)$ denotes the $\varepsilon$-covering number of $\fF|_{\vZ_n}$ w.r.t the $L_\infty$-norm.
			\end{definition}
   \section{Generalization Error in $H^k$ Loss Functions for Supervised Learning}\label{losshk}
In this section, our primary emphasis is on the generalization error of NNs utilizing Sobolev losses with orders $H^k$ where $k$ takes values of 0, 1, and 2. Noting that when $k$ equals 0, the Sobolev loss reduces to $L^2$ loss. As highlighted in the introduction, Sobolev training has found widespread application across numerous tasks such as \cite{adler2018banach, gu2014towards, finlay2018lipschitz, mroueh2018sobolev,finlay2018lipschitz, werbos1992approximate}. 
These specialized loss functions empower models to learn DeNNs capable of approximating the target function with minimal discrepancies in both magnitudes and derivatives.

All the proofs for the content discussed in this section are provided in the appendix. Here, we delve into the generalization error of NNs with $H^k$ loss functions for values of $k$ ranging from $0$ to $2$. Although the proofs in the appendix address each case individually due to differences in their respective formulations, in this section, we consolidate our discussion to encompass all the cases jointly.

\subsection{Notations in Sobolev training}

In a typical supervised learning algorithm, the objective is to learn a high-dimensional target function $f(\vx)$ defined on $[0,1]^d$ with $\|f\|_{W^{n,\infty}\left([0,1]^d\right)}\le 1$ from a finite set of data samples $\{(\vx_i,f(\vx_i))\}_{i=1}^M$. When training a DeNN, we aim to identify a DeNN $\phi$ that approximates $f(\vx)$ based on random data samples $\{(\vx_i,f(\vx_i))\}_{i=1}^M$. We assume that $\{\vx_i\}_{i=1}^M$ is an i.i.d. sequence of random variables uniformly distributed on $[0,1]^d$ in this section. For $H^k$ loss functions for $k=0,1$, we investigate the hypothesis spaces defined as follows:
\begin{align}
\fF_{B,0}=\{&\phi: \text{$\sigma_1$-NNs with width $\le C_1 \log L$,}\notag\\&\text{ depth $\le C_2 L\log L$, $\|\phi\|_{L^{\infty}([0,1]^d)}\le B$}\}\notag\\\fF_{B,1}=\{&\phi: \phi\in\fF_{B,0},~\|\phi\|_{W^{1,\infty}([0,1]^d)}\le B\},\label{f0}
\end{align} where $L\in\sN_+$ and $C_i,B$ are constant will be defined later in Proposition \ref{lul2}.
However, the limitation arises when relying solely on ReLU neural networks, as they lack higher-order derivatives. Consequently, we explore an alternative neural network architecture that extends beyond traditional ReLU networks. This concept, elucidated in \cite{yang2023nearly,yang2023nearlys}, is termed Deep Super ReLU Networks (DSRNs).
Define a subset of $\sigma_2$-NNs characterized by $L\gg 1$ and $C=\fO(1)$ concerning $L$ as follows:
\begin{align}
&\fN_{C,L}:=\{\psi(\vx)=\psi_2\circ\vpsi_1 (\vx): \psi_2\text{ is a $\sigma_2$-NN with depth }\notag\\&\text{$L_2$, each component of $\vpsi_1$ is a $\sigma_1$-NN with depth $L_1$ ,}\notag\\&L_1+L_2\le L,~ L_2\le C \log L.\}
\end{align} 
Let us denote the set as
\begin{align}
\fF_{B,2}=\{\phi: \text{$\phi\in\fN_{6,10(L+1)\log_24L}$, $\|\phi\|_{W^{2,\infty}([0,1]^d)}\le B$}\},\notag
\end{align} where the constants in $\fN_{6,10(L+1)\log_24L}$ are chosen to match the approximation results in \cite{yang2023nearlys}, and we cite the results in the appendix.

Next, let's proceed to define the loss functions in Sobolev training. We denote \begin{align}\fR_{D,k}(\vtheta)&:=\int_{[0,1]^d}h_k(\vx;\vtheta)\,\D \vx,~\fR_{S,k}(\vtheta):=\sum_{i=1}^M \frac{h_k(\vx_i;\vtheta)}{M}.\label{thetaS}
	\end{align}Here, 
$h_k(\vx;\vtheta)$ is defined as follows: \begin{align}
	    h_0(\vx;\vtheta)&:=|f(\vx)-\phi(\vx;\vtheta)|^2,\notag\\h_1(\vx;\vtheta)&:=h_0+|\nabla(f(\vx)-\phi(\vx;\vtheta))|^2,\notag\\h_2(\vx;\vtheta)&:=h_0+|\Delta(f(\vx)-\phi(\vx;\vtheta))|^2\notag
	\end{align}
 For $k=0,1,2$, denote \[\vtheta_{D,k}:=\arg\inf_{\vtheta\in\Theta_k}\fR_{D,k}(\vtheta),~\vtheta_{S,k}:=\arg\inf_{\vtheta\in\Theta_k}\fR_{S,k}(\vtheta)\]with \(\Theta_k:=\{\vtheta\in\sR^W\mid \phi(\vx;\vtheta)\in\fF_{B,k}\}.\)

 The overall inference error (generalization error) is $\rmE\fR_{D,k}(\vtheta_{S,k})$ for $k=0,1,2$, which can be divided into two parts:\begin{align}
		&\rmE\fR_{D,k}(\vtheta_{S,k})\notag\\=& \fR_{D,k}(\vtheta_{D,k})+\rmE\fR_{S,k}(\vtheta_{D,k})-\fR_{D,k}(\vtheta_{D,k})\notag\\+&\rmE(\fR_{S,k}(\vtheta_{S,k})-\fR_{S,k}(\vtheta_{D,k})+\fR_{D,k}(\vtheta_{S,k})-\fR_{S,k}(\vtheta_{S,k}))\notag\\\le &\underbrace{\fR_{D,k}(\vtheta_{D,k})}_{\text{approximation error}}+\underbrace{\rmE\fR_{D,k}(\vtheta_{S,k})-\rmE\fR_{S,k}(\vtheta_{S,k}),}_\text{{sample error}}\label{totl2}
	\end{align}where the last inequality is due to $\rmE\fR_{S,k}(\vtheta_{S,k})\le\rmE\fR_{S,k}(\vtheta_{D,k})$ by the definition of $\vtheta_{S,k}$, and $\rmE\fR_{S,k}(\vtheta_{D,k})-\fR_{D,k}(\vtheta_{D,k})=0$ due to the definition of integration.

 {\begin{remark}
     Our method in the paper can easily generalize to a more general case. If we do not require uniformly distributed input, instead we ask for \(\vx_1,\ldots,\vx_M\sim\mu\) i.i.d. For the sample error, we consider the gap between continuous loss function and discrete loss functions, we just need to change the continuous loss function measured by $\mu(\vx)$. For the sample error part, every analysis is totally the same. For the approximation error, we need to bound this with continuous loss functions. Our result of approximation error is measured in \(W^{k,\infty}\), therefore, the new loss function can also be bounded. Therefore, we can still obtain the optimal generalization error for the new input distribution. For the assumption about $\|f\|_{W^{n,\infty}([0,1]^d)}\le 1$, we can easily generalize it to $\|f\|_{W^{n,\infty}([0,1]^d)}\le D$ for any positive number $D$. Then the approximation error will be multiplied by $D$, and the sample error will not be affected since it only depends on the bound of neural networks $B$. For large $D$, we just need to choose $B \ge D$ in order to contain the neural networks that can approximate the target functions well.
 \end{remark}}

 \subsection{Main results}
In this paper, we explore the generalization error of DeNNs, considering the flexibility in determining the depth of the neural network. 

   \begin{theorem}\label{generalzation l2}
				Let $d,L,M\in\sN_+$, $B,C_1,C_2\footnote{Here, the occurrences of $C_1$ and $C_2$ in $\fF_{B,k}$ continue to hold throughout the rest of the propositions and theorems in this paper.}\in\sR_+$. For any $f\in W^{n,\infty}([0,1]^d)$ with $\|f\|_{W^{n,\infty}([0,1]^d)}\le 1$ for $n>k$ and $k=0,1,2$, we have \begin{align}\rmE\fR_{D,k}(\vtheta_{S,k})\le C\left[ \left(\frac{W}{(\log W)^2}\right)^{-\frac{4 (n-k)}{d}}+\frac{W^2}{M}\log M\right]\notag\end{align}where $W=\fO(L(\log L)^3)$ is the number of parameters in DeNNs, $\rmE$ is expected responding to $X$, $X:=\{\vx_1,\ldots,\vx_M\}$ is an independent random variables set uniformly distributed on $[0,1]^d$, and $C$ is independent with $M,L$.
			\end{theorem}


   \begin{remark}
       The expression \(W = \fO(L(\log L)^3)\) arises from the fact that the number of parameters is equal to the depth times the square of the width. Here, based on the definition of \(\mathcal{F}_{B,k}\), the depth is \(\fO(L\log L)\) and the width is \(\fO(\log L)\). Therefore, we have \(W = \fO(L(\log L)^3)\).
   \end{remark}

\begin{remark}
The generalization error is segmented into two components: the approximation error, denoted by \( \left(\frac{W}{(\log W)^2}\right)^{-\frac{4 (n-k)}{d}} \), and the sample error, represented as \( \frac{W^2}{M} \log M \), where \( W \) is the number of parameters in the neural network.
Note that while the approximation error is contingent on \( k \), the sample error remains unaffected by \( k \).
The sample error's nature is rooted in the Rademacher complexity of several sets, specifically involving derivatives or higher-order derivatives of neural networks. In the appendix, our analysis demonstrates that these complexities maintain a consistent order relative to the depth and width of NNs.
Consequently, the constant \( C \) in the expression \( \frac{W^2}{M} \log M \) is dependent on \( k \), while the other terms remain independent of it.
\end{remark}
The outcome is nearly optimal with respect to the number of sample points $M$ \cite{schmidt2020nonparametric} based on following corollary.
\begin{corollary}\label{opt}
    Let \(d, M \in \mathbb{N}_+\), \(B, C_1, C_2 \in \mathbb{R}_+\). For any \(f \in W^{n,\infty}([0,1]^d)\) with \(\|f\|_{W^{n,\infty}([0,1]^d)} \leq 1\) for \(n > k\) and \(k = 0,1,2\), we have
\[
\rmE{\fR_{D,k}}(\boldsymbol{\theta}_{S,k}) \leq C M^{-\frac{2(n-k)}{2(n-k)+d}}
\]
where the result is up to the logarithmic term, \(\rmE\) is expected responding to \(X\), and \(X := \{\vx_1, \ldots, \vx_M\}\) is an independent set of random variables uniformly distributed on \([0,1]^d\). \(C\) is a constant independent of \(M\).
\end{corollary}
\begin{proof}
   Set $W=M^{\frac{d}{2d+4n}}$ in Theorem \ref{generalzation l2}, and the result can be obtained directly.
\end{proof}
\begin{remark}
The result in the corollary is nearly optimal, as indicated by the findings in \cite{schmidt2020nonparametric}.  In other words, we achieve nearly optimal generalization error for DeNNs with any number of hidden layers, and this differs from the results presented in \cite{schmidt2020nonparametric, suzuki2018adaptivity}. {Furthermore, the consideration of Sobolev training within Sobolev spaces indeed introduces challenges due to the curse of dimensionality inherent in such spaces. This issue has been extensively discussed and proven in the paper \citep{yang2023nearly}. It's important to emphasize that addressing the curse of dimensionality within Sobolev spaces is not the focus of our method. Instead, in this paper, we propose a method to determine the optimal generalization error of Sobolev training.

If one aims to achieve generalization results in Sobolev training without encountering the curse of dimensionality, exploring smaller spaces such as Korobov spaces and Barron spaces, which are subspaces of Sobolev spaces, becomes crucial. Our method is fully applicable in these spaces and can accurately determine the generalization error of Sobolev training without being affected by the curse of dimensionality.}
\end{remark}

Based on our analysis, it's important to recognize that DeNNs ($N = \mathcal O(1)$ or ${\mathcal O}\left(\log(L)\right)$) with super convergence approximation rates may not necessarily outperform shallow or less DeNNs. This is because, in very deep NNs lacking proper parameter control boundaries, the hypothesis space can become excessively large, resulting in a significant increase in sample error. Nonetheless, it's crucial to acknowledge that DeNNs still offer advantages. 
Consider a large number $W \gg 1$. Let's assume that both deep and shallow (or not very deep) neural networks have $W$ parameters. The generalization error in DeNNs can be characterized as
\(
\fO\left(W^{-\frac{4(n-k)}{d}}+\frac{W^2}{M}\right),
\)
While the generalization error in shallow (or not very deep) neural networks can be obtained by the results shown in \cite{schmidt2020nonparametric,suzuki2018adaptivity} or based on our method shown in this paper:
\begin{proposition}\label{l2bound}
Let \(d,L,N,M\in\mathbb{N}_+\), \(B,C_1,C_2\in\mathbb{R}_+\). For any \(f\in W^{n,\infty}([0,1]^d)\) with \(\|f\|_{W^{n,\infty}([0,1]^d)}\le 1\) for \(n>k\) and \(k=0,1,2\), we have
\begin{align}
&\rmE\mathcal{R}_{D,k}(\boldsymbol{\theta}_{S,k})\notag\\\le& C\left[ \left(NL\right)^{-\frac{4 (n-k)}{d}}+\frac{N^2L^2\log N\log L}{M}\log M\right],
\end{align}
where \(\rmE\) is the expected value with respect to \(X\), \(X:=\{\vx_1,\ldots,\vx_M\}\) is an independent set of random variables uniformly distributed on \([0,1]^d\), and \(C\) is independent of \(M,L,N\).
\end{proposition}

Based on the result in Proposition \ref{l2bound}, we know that the generalization error up to a logarithmic factor of WeNNs ($L = \mathcal O(1)$ or ${\mathcal O}\left(\log(N)\right)$) is
\(
\fO\left(W^{-\frac{2(n-k)}{d}}+\frac{W}{M}\right)
\) based on the estimation of covering number shown in \cite{suzuki2018adaptivity}.
When \(M \ge W^{\frac{2n+2d-2k}{d}}\), the order of the generalization error in DeNNs surpasses that of WeNNs. This relationship is illustrated in Fig.~\ref{deep_shallowall}. This tells us that for a fixed number of sample points, if you want your networks to have a few parameters, DeNNs are a better choice; otherwise, opt for a shallow neural network. If the number of parameters is nearly fixed from the start and the sample size is very large, consider using DeNNs; otherwise, a shallow network is better to control the generalization error.

\begin{remark}
   Our analysis in the paper focuses on \( k = 0, 1, 2 \), and it is straightforward to generalize to \( k \geq 3 \). Regarding the approximation result, reference can be made to \cite{yang2023nearlys,guhring2021approximation}. For the generalization error, the estimation method of \( H^2 \) loss functions in this paper can be directly extended to estimate the sample error in \( H^k \) loss functions.
\end{remark}

Another finding based on our analysis is that, as the value of \(k\) increases, the curve will shift to the left in Fig.~\ref{deep_shallowall}, expanding the region of better performance for DeNNs. This discovery suggests that if the loss function requires higher regularity, DeNNs may be a preferable choice over wider neural networks. In this paper, we consider the underparameterized case, while the overparameterized cases are regarded as future work.



{\subsection{Proof Sketche of Theorem \ref{generalzation l2}}
In this subsection, we present the proof sketches for estimating the generalization error of DeNNs in Sobolev training and compare them with the differences observed in shallow neural networks with $L^2$-loss.

$\bullet$ First and foremost, we apply the inequality established in \cite{gyorfi2002distribution}, as presented in Lemma \ref{connect}, which utilizes uniform covering numbers (refer to Definition \ref{uniform}) to bound the generalization errors. This method allows us to obtain the optimal generalization error with respect to the number of samples. Additionally, in Sect. \ref{improve}, we introduce another approach using the Rademacher complexity to bound the generalization error. While this method may yield suboptimal results in terms of the number of sample points, it proves to be more friendly to the number of parameters, making it suitable for overparameter analysis. This is regarded as future work.

$\bullet$ Secondly, we employ pseudo-dimension (see Definition \ref{Pse}) to bound the uniform covering number.

The outcomes of these first two steps differ from those in shallow or not very deep neural networks, as presented in \cite{guhring2020error, schmidt2020nonparametric, suzuki2018adaptivity}. In those studies, a universal bound on parameters was established for all functions. In such instances, the complexity of neural networks could be effectively regulated using covering numbers instead of uniform covering numbers. For covering numbers with a uniform bound parameter in neural networks, estimation can be performed directly, even if they contain derivatives. We present the proof in the appendix.

However, in the approximation of DeNNs, some parameters may become large to approximate specific functions due to the use of techniques called bit extraction in the approximation. These techniques enable better approximation results, as demonstrated in \cite{lu2021deep, siegel2022optimal}. Therefore, we cannot assume a uniform bound like in shallow neural networks. This makes the estimation more challenging, and this is the result we will use uniform covering numbers to bound the generalization errors as Lemma \ref{connect} and then use the pseudo-dimension to bound the uniform covering number.

$\bullet$ Last of all, we estimate the pseudo-dimension of DeNNs and functions represented as higher-order differential operator actions on DeNNs. The structure of derivative and higher-order derivative DeNNs is very complex, involving tensor products and the addition of many relative parts due to the chain rule, which makes estimating the richness or complexity of such spaces difficult.}

Our theoretical results can also be validated through experimentation Table \ref{table2}. In the appendix (see Appendix \ref{experiment}), we provide a detailed account of a simple experiment conducted to corroborate our findings.

\begin{table*}[h]\label{table2}
    \centering
    \begin{tabular}{|c|c|c|}
    \hline
       Neural Network  &  Large Data Regime & Small Data Regime \\
       \hline
       Shallow (Depth 1, Width 20)  & Mean: 6.18e-4, Std: 9.86e-05 & Mean: 1.51e-3, Std: 2.52e-4 \\
       \hline
       Deep (Depth 4, Width 10)      & Mean: 3.69e-4, Std: 2.57e-05 & Mean: 4.96e-3, Std: 4.46e-3 \\
       \hline
    \end{tabular}
    \caption{This table compares the performance of shallow and deep neural networks in terms of mean test performance and standard deviation across different data regimes. It illustrates how network depth and data availability impact learning outcomes, with shallow networks performing better in small data scenarios, while deep networks excel with larger datasets.}
\end{table*}

\section{Applications for Solving Partial Differential Equations}\label{apply} One of the most crucial applications of neural networks is in solving partial differential equations (PDEs), exemplified in methods such as deep Ritz \cite{weinan2017deep}, PINN \cite{raissi2019physics}, and deep Galerkin method (DGM) \cite{sirignano2018dgm}. These methods construct solutions using neural networks, designing specific loss functions and learning parameters to approximate the solutions. The choice of loss function varies among methods; for instance, deep Ritz methods use the energy variation formula, while PINN and DGM employ residual error.

In these approaches, loss functions integrate orders of, or higher-order derivatives of, neural networks. This design allows models to learn neural networks capable of approximating the target function with minimal discrepancies in both magnitude and derivative. We specifically focuses on solving partial differential equations using neural networks by directly learning the solution. Another approach, known as operator learning, involves learning the mapping between input functions and solutions, but this method is not discussed in this paper. In this section, we delve into the Poisson equation solved using both the deep Ritz method \cite{weinan2017deep} and the PINN method \cite{raissi2019physics}:
\begin{equation}
\begin{cases}
-\Delta u = f & \text { in } \widehat{\Omega}, \\
u = 0 ~\text{or}~\frac{\partial u}{\partial \nu}=0 & \text { on } \partial \widehat{\Omega},
\end{cases}
\label{PDE}
\end{equation}
where $f\in L^2(\widehat{\Omega})$. The domain $\widehat{\Omega}$ is defined in $[0,1]^d$ with a smooth boundary. The assumption of a smooth and bounded $\widehat{\Omega}$ is made to establish a connection between the regularity of the boundary and the interior, facilitated by trace inequalities. The proofs of theorem in this section are presents in the appendix.
   \subsection{Generalization Error in deep Ritz Method} 
   
   We initially focus on solving the Poisson equation, as represented by Eq.~(\ref{PDE}) with Neumann boundary condition, using the deep Ritz method. This approach utilizes the energy as the loss function during training to guide the neural networks in approximating the solutions. In the context of the Poisson equation, the energy functional encompasses the first derivative of the neural networks. Consequently, based on \cite{lu2021priori}, the corresponding loss function in deep Ritz methods can be expressed as:
\begin{align}
\fE_{D,r}(\phi) :=&  \int_{\widehat{\Omega}} g(\vx;\vtheta) \, \D \vx + \left(\int_{\widehat{\Omega}} \phi(\vx;\vtheta) \, \D \vx\right)^2,\notag\\g(\vx;\vtheta):=&\frac{1}{2}|\nabla \phi(\vx;\vtheta)|^2 - f\phi(\vx;\vtheta),
\notag\end{align}
where $\vtheta$ represents all the parameters in the neural network.
To discretize the loss function, we randomly choose $M_1$ points in the interior and $M_2$ points on the boundary:
\begin{align}
    \fE_{S,r}(\phi) :=& \frac{1}{2M_1} \sum_{i=1}^{M_1} g(\vx_i;\vtheta)+ \left(\frac{1}{M_2}\sum_{i=1}^{M_2} \phi(\vy_i;\vtheta)\right)^2.\notag
\end{align}
Denote \(\vtheta_{D,r}:=\arg\inf_{\vtheta\in\Theta_1} \fE_{D,r}(\phi(\vx;\vtheta)),~\vtheta_{S,r}:=\arg \inf_{\vtheta\in\Theta_1} \fE_{S,r}(\phi(\vx;\vtheta))\).

In the deep Ritz method, a well-learned solution is characterized by the smallness of the error term $\rmE\fE_{D,r}(\phi(\vx;\vtheta_{S,r}))-\fE_{D,r}(u^*)$, where $u^*(\vx)$ represents the exact solution of Eq.~(\ref{PDE}). This error can be decomposed into two parts:
\begin{align}
&\rmE\fE_{D,r}(\phi(\vx;\vtheta_{S,r}))-\fE_{D,r}(u^*)\notag\\
\le &\underbrace{\rmE\fE_{D,r}(\phi(\vx;\vtheta_{S,r}))-\rmE\fE_{S,r}(\phi(\vx;\vtheta_{S,r}))}_{\text{sample error}}\notag\\
&+\underbrace{\fE_{D,r}(\phi(\vx;\vtheta_{D,r}))-\fE_{D,r}(u^*)}_{\text{approximation error}}.
\end{align}
For the approximation error, it can be controlled by the $H^1$ gap between $\phi(\vx;\vtheta_{D,r})$ and $u^*$ based on the trace inequality.
\begin{lemma}\label{app r}
    Suppose that $\phi\in \fF_{B,1}$ and $u^*(\vx)$ is the exact solution of Eq.~(\ref{PDE}) with Neumann boundary condition, then there is a constant $C$ such that \(\fE_{D,r}(\phi)-\fE_{D,r}(u^*)\le C\|\phi-u^*\|^2_{H^1(\widehat{\Omega})}.\)
\end{lemma}
\begin{proof}
    The proof can be found in \cite{lu2021priori}.
\end{proof}
Regarding the sample error, it can be estimated using the results in Theorem \ref{generalzation l2} for $k=1$. Therefore, by combining the estimations of the sample error and the approximation error, we obtain the following theorem:
\begin{theorem}\label{generalzation r}
				Let $d, L,M_1,M_2\in\sN_+$, $B,C_1,C_2\in\sR_+$. For any $f\in W^{n,\infty}([0,1]^d)$ with $\|f\|_{W^{n,\infty}([0,1]^d)}\le 1$, we have an independent constant $C$ such that \begin{align}&\rmE\fE_{D,r}(\phi(\vx;\vtheta_{S,r}))-\fE_{D,r}(u^*)\notag\\\le& C\left[ \left(\frac{W}{(\log W)^2}\right)^{-\frac{4 (n-1) }{ d}}+W^2\left(\frac{\log M_1}{{M_1}}+\frac{\log M_2}{{M_2}}\right)\right] \notag\end{align}where $W=\fO(L(\log L)^3)$ is the number of parameters in DeNNs, $u^*$ is the exact solution of Eq.~(\ref{PDE}), $\rmE$ is expected responding to $X$ and $Y$, and $X:=\{\vx_1,\ldots,\vx_{M_1}\}$ and $Y:=\{\vy_1,\ldots,\vy_{M_2}\}$ is an independent random variables set uniformly distributed on $\widehat{\Omega}$ and $\partial \widehat{\Omega}$.
			\end{theorem}

   The error is decomposed into three components. The term $\left(\frac{W}{(\log W)^2}\right)^{-\frac{4 (n-1) }{ d}}$ represents the optimal approximation error, while 
   $\frac{\log M_i}{{M_i}}$ corresponds to the sample error.

\subsection{Generalization error of Physics-Informed Neural Network Methods}
 In this subsection, we apply PINN methods to solve the Poisson equation Eq.~\eqref{PDE} with Dirichlet boundary condition. PINN differs from the deep Ritz method in that it uses the residual error as the loss function, which is an $H^2$ loss function, similar to the DGM. Therefore, we will leverage the result from Theorem \ref{generalzation l2} for the case where $k=2$.
The corresponding loss function in Physics-Informed Neural Network methods can be expressed as:
\begin{align}
\fE_{D,p}(\phi) &:=  \int_{\widehat{\Omega}} |\Delta \phi(\vx;\vtheta)+f(\vx)|^2 \, \D \vx + \lambda \int_{\partial \widehat{\Omega}} \phi(\vx;\vtheta)^2 \, \D s.
\notag\\
    \fE_{S,p}(\phi) &:= \sum_{i=1}^{M_1} \frac{|\Delta\phi(\vx_i;\vtheta)+f(\vx)|^2}{M_1} + \lambda\sum_{i=1}^{M_2} \frac{\phi^2(\vy_i;\vtheta)}{M_2}.\notag
\end{align}
Denote \(\vtheta_{D,p}:=\arg\inf_{\vtheta\in\Theta_2} \fE_{D,p}(\phi(\vx;\vtheta)),~\vtheta_{S,p}:=\arg \inf_{\vtheta\in\Theta_2} \fE_{S,p}(\phi(\vx;\vtheta)).
\)
In the PINN method, if the solution is learned well, meaning that $\rmE\fE_{D,p}(\phi(\vx;\vtheta_{S,p}))$ should be small. It still can be divided into two parts, approximation error and sample error. The approximation error can be read as \(\fE_{D,p}(\phi(\vx;\vtheta_{D,p}))\) and sample error can be read as \(\rmE\fE_{D,p}(\phi(\vx;\vtheta_{S,p}))-\rmE\fE_{S,p}(\phi(\vx;\vtheta_{S,p}))\). The approximation error can be bounded as the following lemma:\begin{lemma}\label{H2}
     Suppose that $\phi(\vx;\vtheta_{D,p})\in \fF_{B,2}$ and $u^*$ is the exact solution of Eq.~(\ref{PDE}) with Dirichlet boundary condition, then there is a constant $C$ such that \(\fE_{D,p}(\phi(\vx;\vtheta_{D,p}))\le C\|\phi(\vx;\vtheta_{D,p})-u^*\|^2_{H^2(\widehat{\Omega})}.\)
\end{lemma}By applying Theorem \ref{generalzation l2} and above lemma, we can estimate overall inference error for DeNNs .
\begin{theorem}\label{generalzation p}
				Let $d, L,M_1,M_2\in\sN_+$, $B,C_1,C_2\in\sR_+$. For any $f\in W^{n,\infty}([0,1]^d)$ with $\|f\|_{W^{n,\infty}([0,1]^d)}\le 1$, we have an independent constant $C_{9}$ such that \begin{align}&\rmE\fE_{D,p}(\vtheta_{S,p})\le C\big[ \left({W}/{(\log W)^2}\right)^{-\frac{4 (n-2) }{ d}}\notag\\&+W^2\left({\log M_1}/{{M_1}}+{\log M_2}/{{M_2}}\right)\big]\notag\end{align}where $W=\fO(L(\log L)^3)$ is the number of parameters in DeNNs, 
    $C$ is a constant independent with $M,L$.
			\end{theorem}

   The performance comparison between shallow NNs and DeNNs in supervised learning (Section \ref{losshk}) can be directly applied here. There is no need to reiterate the discussion in this context.

   {\begin{remark}
       In this section, we only consider Laplace equations. However, our method can work in a more general case. The difference between Laplace equations and general PDEs lies in two aspects. Firstly, we need to utilize a PDE method to bound the approximation error in methods like deep Ritz or PINN by the Sobolev norm of the gap between the neural network and the exact solution. This step only requires assuming that the coefficients in the PDEs are smooth and bounded. Secondly, in the sample error, we need to consider the VC-dimension of the neural network acting as an operator in the PDEs. The approach to this is similar to what we have presented in our paper. Hence, our method works for general PDEs with some assumptions on the coefficients of the PDEs.
   \end{remark}}

   \section{Improving Generalization Error Across Parameter Counts}\label{improve}
   Another approach to bounding the generalization error is based on the Rademacher complexity. The result differs from Theorem \ref{generalzation l2}, and it provides a better outcome with respect to the number of parameters. However, it is not nearly optimal in terms of the number of sample points. The detailed proof is provided in the appendix. In the subsequent comparison between WeNNs and DeNNs, we will utilize Theorem \ref{generalzation l2} instead of this result, as this paragraph focuses on the underparameterized cases. For the overparameterized cases, estimation of Rademacher complexity may prove useful.

 \begin{theorem}\label{generalzation l2R}
				Let $d, L,M\in\sN_+$, $B,C_1,C_2\in\sR_+$. For any $f\in W^{n,\infty}([0,1]^d)$ with $\|f\|_{W^{n,\infty}([0,1]^d)}\le 1$ for $n>k$ and $k=0,1,2$, we have \begin{align}&\rmE\fR_{D,k}(\vtheta_{S,k})\le C\left[\left(\frac{W}{(\log W)^2}\right)^{-\frac{4 (n-k)}{d}}+\frac{W}{\sqrt{M}}\log M\right]\notag\end{align}where $W=\fO(L(\log L)^3)$ is the number of parameters in DeNNs, $\rmE$ is expected responding to $X$, $X:=\{\vx_1,\ldots,\vx_M\}$ is an independent random variables set uniformly distributed on $[0,1]^d$, and $C$ is independent with $M,L$.
			\end{theorem}

   \section{Conclusions and Future Works}
   In this paper, we examine the optimal generalization error in strategies involving DeNNs versus WeNNs. Our finding suggests that when the parameter count in neural networks remains constant, the decision between WeNNs and DeNNs hinges on the volume of available sample points. On the other hand, when the sample points are predetermined, the choice between WeNNs and DeNNs is contingent upon specific objective functions. Furthermore, our analysis underscores the pivotal role of the loss function's regularity in determining neural network stability. Specifically, in the context of Sobolev training, when the derivative order within the loss function is elevated, favoring DeNNs over their shallow counterparts emerges as a prudent strategy.

We specifically compare DeNNs, characterized by an arbitrary number of hidden layers, in the under-parameterized case. The overparameterized scenario is considered a topic for future research. Additionally, the extension to more complex spaces, such as Korobov spaces \cite{montanelli2019new,yang2023optimal} and functional spaces \cite{yang2022approximation}, is envisaged as part of our future research endeavors. Moreover, in light of the recently established approximation theory for convolutional neural networks (CNNs) \cite{zhou2020universality,he2022approximation}, we plan to investigate the generalization error estimates for CNNs.

\section{Acknowledgments and Statements}
We extend our heartfelt gratitude to Prof. Wenrui Hao and Dr. Xinliang Liu for their invaluable contributions to discussions, experiments, and feedback, which greatly enhanced the quality of this work.
The work of the second author was supported by the KAUST Baseline Research Fund.

This paper represents an effort to push the boundaries of the field of Machine Learning. While our work may have various potential societal implications, we do not find it necessary to highlight any specific ones at this time.
\bibliography{references}
\bibliographystyle{plainnat}

\newpage
\appendix
\onecolumn
\section{Approximation Results for DeNNs with Super Convergence Rates}
To bound the approximation error in Eq.~(\ref{totl2}) for $k=0$, we employ the following nearly optimal approximation result, which can be found in \cite{lu2021deep, siegel2022optimal}:\begin{proposition}[\cite{lu2021deep, siegel2022optimal}]\label{lul2}
      Given a function $f(\vx)$ defined on $[0,1]^d$ with $\|f\|_{W^{n,\infty}\left([0,1]^d\right)}\le 1$, for any $N, L \in \sN^{+}$, there exists a constant $B\ge 1$ and a function implemented by a $\sigma_1$-NN with $C_1 N$ width and $C_2 L$ depth such that
$$
\|\phi-f\|_{L^{\infty}\left([0,1]^d\right)} \leq C_{3,0} N^{-2 n / d} L^{-2 n / d},
$$
where $C_1, C_2$ and $C_{3,0}$ are constants independent with $N$ and $L$.
\end{proposition}
To bound the approximation error in Eq.~(\ref{totl2}) for $k=1$, we employ the following nearly optimal approximation result, which can be found in \cite{yang2023nearly}
 \begin{proposition}[{\cite{yang2023nearly}}]\label{yangh1}
      Given a function $f(\vx)$ defined on $[0,1]^d$ with $\|f\|_{W^{n,\infty}\left([0,1]^d\right)}\le 1$, for any $N, L \in \sN^{+}$, there exists a constant $B\ge 1$ and a function implemented by a $\sigma_1$-NN with $C_1 N$ width and $C_2 L$ depth such that
$$
\|\phi-f\|_{W^{1,\infty}\left([0,1]^d\right)} \leq C_{3,1} N^{-2 (n-1) / d} L^{-2 (n-1) / d},
$$
where $C_1, C_2$ and $C_{3,1}$ are constants independent with $N$ and $L$.
 \end{proposition}

 For the approximation error n Eq.~(\ref{totl2}) for $k=2$, we can directly leverage the findings presented in \cite{yang2023nearlys}:
 \begin{proposition}
 \label{main2}[{\cite{yang2023nearlys}}]
	For any $f\in W^{n,p}((0,1)^d)$ with $\|f\|_{W^{n,p}((0,1)^d)}\le 1$ for $p\in[1,\infty]$, any $d,n\ge2$ and $N, L\in\sN_+$ with $N \log_2L+2^{\left\lfloor\log _2 N\right\rfloor} \geq \max\{d,n\}$ and $L\ge N$, there is a DSRN $\phi(\vx)$ in $\fN_{6,10(L+1)\log_24L}$ with the width $2^{d+6}n^{d+1}(N+d)\log_2(8N)$ such that$$
\|\phi-f\|_{W^{2,\infty}\left([0,1]^d\right)} \leq C_{3,2} N^{-2 (n-2) / d} L^{-2 (n-2) / d},
$$where $C_{3,2}$ is the constant independent with $N,L$.
\end{proposition}

\section{Lemmas Related to Covering Numbers}
The following lemma will be used to bounded generalization error by covering numbers:
			
			\begin{lemma}[\cite{gyorfi2002distribution}, Theorem 11.4]\label{connect}
				Let $M \in \sN$, and assume $\|f\|_{L^{\infty}(\Omega)}\le B$ for some $B \geq 1$. Let $\fF$ be a set of functions from $\fX$ to $[-B, B]$. Then for any $0<\epsilon \leq 1 / 2$ and $\alpha, \beta>0$,
$$
\begin{aligned}
& \rmP\left\{\exists \phi \in \fF:\left\|f-\phi\right\|_{L^2}^2-\left(\fR_{S,0}(\phi)-\fR_{S,0}\left(f\right)\right) \geq \epsilon\left(\alpha+\beta+\left\|f-\phi\right\|_{L^2}^2\right)\right\} \\
& \leq 14 \fN\left(\frac{\beta \epsilon}{20 B}, \fF, M\right) \exp \left(-\frac{\epsilon^2(1-\epsilon) \alpha M}{214(1+\epsilon) B^4}\right),
\end{aligned}
$$where \[\fR_{S,0}(g)=\frac{1}{M}\sum_{i=1}^M|g(\vx_i)-y_i|^2\] for $\{y_i\}_{i=1}^M\subset [-B,B]$ and $\{\vx_i\}_{i=1}^M\subset \Omega$.
			\end{lemma}
			
			We estimate the uniform covering number by the pseudo-dimension based on the following lemma.

   \begin{definition}
   [pseudo-dimension \citep{pollard1990empirical}]\label{Pse}
		Let $\fF$ be a class of functions from $\fX$ to $\sR$. The pseudo-dimension of $\fF$, denoted by $\text{Pdim}(\fF)$, is the largest integer $m$ for which there exists $(x_1,x_2,\ldots,x_m,y_1,y_2,\ldots,y_m)\in\fX^m\times \sR^m$ such that for any $(b_1,\ldots,b_m)\in\{0,1\}^m$ there is $f\in\fF$ such that $\forall i: f\left(x_i\right)>y_i \Longleftrightarrow b_i=1.$    
   \end{definition}

   \begin{lemma}[\cite{anthony1999neural}]\label{cover dim}
				Let $\fF$ be a class of functions from $\fX$ to $[-B,B]$. For any $\varepsilon>0$, we have \[\fN(\varepsilon,\fF,n)\le \left(\frac{2enB}{\varepsilon\text{Pdim}(\fF)}\right)^{\text{Pdim}(\fF)}\] for $n\ge \text{Pdim}(\fF)$.
			\end{lemma}

 \section{Proof of Theorem \ref{generalzation l2}}
There are three cases in Theorem \ref{generalzation l2}. In order to prove Theorem \ref{generalzation l2}, we consider these three cases individually. \begin{remark}
    For simplicity in notation, we denote the width as \(N = \fO(\log L)\) in \(\mathcal{F}_{B,k}\) in the following proofs. In order to obtain the result for DeNNs, we just substitute \(N = \fO(\log L)\) in the final result.

\end{remark}

\subsection{Proof of Theorem \ref{generalzation l2} for $k=0$.}
		
			 We first show the proof of the Theorem \ref{generalzation l2} in $k=0$ case:
    			\begin{proposition}
			\label{connect rad l2}
				Let $d, N, L,M\in\sN_+$, $B,C_1,C_2\footnote{Here, the occurrences of $C_1$ and $C_2$ in $\fB_{B,0}$ continue to hold throughout the rest of the propositions and theorems in this paper.}\in\sR_+$. For any $f\in W^{n,\infty}([0,1]^d)$ with $\|f\|_{W^{n,\infty}([0,1]^d)}\le 1$, we have \begin{align}&\rmE\fR_{D,0}(\vtheta_{S,0})\le \frac{5136 B^4}{M}\left[\log\left(14 \fN\left(\frac{ 1}{80 BM}, \fF_{B,0}, M\right) \right)+1\right]+2\inf_{\phi\in\fF_{B,0}}\|\phi-f\|_{L^{\infty}\left([0,1]^d\right)},\notag\end{align}where \(\fF^2_{B,0}:=\{\bar{\phi}: \bar{\phi}=\phi^2,\phi\in\fF_{B,0}\},\) $\rmE$ is expected responding to $X$, and $X:=\{\vx_1,\ldots,\vx_M\}$ is an independent random variables set uniformly distributed on $[0,1]^d$.
			\end{proposition}
\begin{proof}
 Due to the definition, we know that \[\fR_{D,0}(\vtheta_{D,0})=\inf_{\phi\in\fF_{B,0}}\|\phi-f\|_{L^{2}\left([0,1]^d\right)}.\] For the sample error, due to $\vtheta_{S,0}$ and $\vtheta_{D,0}$ belong to $\Theta$ almost surely, we have\begin{align}
					&  \rmE\left[\fR_{D,0}(\vtheta_{S,0})-\fR_{S,0}(\vtheta_{S,0})\right]\notag\\=&\rmE\left[\int_{[0,1]^d}\left| f(\vx)-\phi(\vx;\vtheta_{S,0})\right|^2\,\D \vx-\frac{1}{M}\sum_{i=1}^M\left| f(\vx_i)-\phi(\vx_i;\vtheta_{S,0})\right|^2\right]\notag\\=&\rmE\left[\int_{[0,1]^d}\left| f(\vx)-\phi(\vx;\vtheta_{S,0})\right|^2\,\D \vx-\frac{2}{M}\sum_{i=1}^M\left| f(\vx_i)-\phi(\vx_i;\vtheta_{S,0})\right|^2\right]+\rmE\frac{1}{M}\sum_{i=1}^M\left| f(\vx_i)-\phi(\vx_i;\vtheta_{S,0})\right|^2\notag.
				\end{align}
Based on the definition of $\vtheta_{S,0}$, we know that \begin{align}
    \rmE\frac{1}{M}\sum_{i=1}^M\left| f(\vx_i)-\phi(\vx_i;\vtheta_{S,0})\right|^2\le \inf_{\phi\in\fF_{B,0}}\|\phi-f\|_{L^{\infty}\left([0,1]^d\right)}
\end{align} Next we denote \begin{equation}
    \rmA_0:=\int_{[0,1]^d}\left| f(\vx)-\phi(\vx;\vtheta_{S,0})\right|^2\,\D \vx-\frac{2}{M}\sum_{i=1}^M\left| f(\vx_i)-\phi(\vx_i;\vtheta_{S,0})\right|^2.\label{a0}\end{equation} Note that \(\fR_{S,0}\left(f\right)=0\) and\begin{align}
    \rmP(\rmA_0\ge \epsilon)&=\rmP\left( 2\int_{[0,1]^d}\left| f(\vx)-\phi(\vx;\vtheta_{S,0})\right|^2\,\D \vx-\frac{2}{M}\sum_{i=1}^M\left| f(\vx_i)-\phi(\vx_i;\vtheta_{S,0})\right|^2\ge\epsilon+\int_{[0,1]^d}\left| f(\vx)-\phi(\vx;\vtheta_{S,0})\right|^2\,\D \vx\right)\notag\\&=\rmP\left( 2\left(\|f-\phi\|_{L^2([0,1]^d)}-\left(\fR_{S,0}(\phi)-\fR_{S,0}\left(f\right)\right)\right)\ge\epsilon+\|f-\phi\|_{L^2([0,1]^d)}\right)\notag\\&=\rmP\left( \|f-\phi\|_{L^2([0,1]^d)}-\left(\fR_{S,0}(\phi)-\fR_{S,0}\left(f\right)\right)\ge\frac{1}{2}\left(\frac{1}{2}\epsilon+\frac{1}{2}\epsilon+\|f-\phi\|_{L^2([0,1]^d)}\right)\right)\notag\\&\le 14 \fN\left(\frac{ \epsilon}{80 B}, \fF_{B,0}, M\right) \exp \left(-\frac{\epsilon M}{5136 B^4}\right),
\end{align}where the last inequality is due to Lemma \ref{connect}.

Therefore, we have that \begin{align}
    \rmE \rmA_0&\le \int_{0}^\infty \rmP(\rmA_0\ge y)\,\D y\le\epsilon+\int_{\epsilon}^\infty \rmP(\rmA_0\ge y)\,\D y\notag\\&\le\epsilon+\int_{\epsilon}^\infty 14 \fN\left(\frac{ \epsilon}{80 B}, \fF_{B,0}, M\right) \exp \left(-\frac{y M}{5136 B^4}\right)\,\D y
\end{align}

By the direct calculation, we have \begin{align}
    \int_{\epsilon}^\infty 14 \fN\left(\frac{ \epsilon}{80 B}, \fF_{B,0}, M\right) \exp \left(-\frac{y M}{5136 B^4}\right)\,\D y\le 14 \fN\left(\frac{ \epsilon}{80 B}, \fF_{B,0}, M\right) \frac{5136 B^4}{M}\exp \left(-\frac{\epsilon M}{5136 B^4}\right).
\end{align} Set \[\epsilon=\frac{5136 B^4}{M}\log\left(14 \fN\left(\frac{ 1}{80 BM}, \fF_{B,0}, M\right) \right)\ge \frac{1}{M}\] and we have \[\rmE \rmA_0\le \frac{5136 B^4}{M}\left[\log\left(14 \fN\left(\frac{ 1}{80 BM}, \fF_{B,0}, M\right) \right)+1\right]\]

Hence we have \begin{align}
        \rmE\left[\fR_{D,0}(\vtheta_{S,0})-\fR_{S,0}(\vtheta_{S,0})\right]\le \frac{5136 B^4}{M}\left[\log\left(14 \fN\left(\frac{ 1}{80 BM}, \fF_{B,0}, M\right) \right)+1\right]+\inf_{\phi\in\fF_{B,0}}\|\phi-f\|_{L^{\infty}\left([0,1]^d\right)}.
    \end{align}
			\end{proof}

Next we need to bound the covering number $\fN\left(\frac{ 1}{80 BM}, \fF_{B,0}, M\right)$, which can be estimate by the $\text{Pdim}(\fF_{B,0})$ based on Lemma \ref{cover dim}. Based on \cite{bartlett2019nearly}, $\text{Pdim}(\fF_{B,0})=\fO(L^2N^2\log_2 L\log_2 N)$:

   \begin{proposition}[\cite{bartlett2019nearly}]\label{pdim1}
			For any $N,L,d\in\sN_+$, there exists a constant $\widehat{C}$ independent with $N,L$  such that 	\begin{equation}
				\text{Pdim}(\fF_{B,0})\le \widehat{C} N^2L^2\log_2 L\log_2 N.
			\end{equation}
		\end{proposition}

  Now we can show the proof of Theorem \ref{generalzation l2} for $k=0$:\begin{proof}[Proof of Theorem \ref{generalzation l2} for $k=0$]
				 Let $J=\text{Pdim}(\fF_{B,0})$. Due to Lemma \ref{cover dim}, for any $M\ge J$, we have \begin{align}
					\log\left(14 \fN\left(\frac{ 1}{80 BM}, \fF_{B,0}, M\right) \right)\le& \log\left(14 \left(\frac{160eM^2B^2}{\text{Pdim}(\fF_{B,0})}\right)^{\text{Pdim}(\fF_{B,0})} \right)\notag\\\le &C \text{Pdim}(\fF_{B,0})\log M &\notag\\\le&C\widehat{C} N^2L^2\log_2 L\log_2 N\log M.
				\end{align}
				
				Hence, we have that there is a constant $C_5=C_5(B,d, C_1,C_2):=10272 B^4C\widehat{C}$ such that\begin{equation}
					\rmE\fR_{D,0}(\vtheta_{S,0})\le C_{3,0} N^{-2 n / d} L^{-2 n / d}+C_5\frac{N^2L^2\log_2 L\log_2 N}{M}\log M.
				\end{equation}

    Finally, we set \(N = \fO(\log L)\) according to the definition of DeNNs, and the number of parameters \(W = \fO(N^2L\log L) = \fO(L(\log L)^3)\) to complete the proof.
			\end{proof}

   \subsection{Proof of Theorem \ref{generalzation l2} for $k=1$}
  The Theorem \ref{generalzation l2} in the \(k=1\) case can be read as follows:

 \begin{proposition}
			\label{connect radh1}
				Let $d, N, L,M\in\sN_+$, $B,C_1,C_2\in\sR_+$. For any $f\in W^{n,\infty}([0,1]^d)$ with $\|f\|_{W^{n,\infty}([0,1]^d)}\le 1$, we have \begin{align}
        &\rmE\fR_{D,1}(\vtheta_{S,1})\le \frac{5136 B^4}{M}\left[\log\left(14 \fN\left(\frac{ 1}{80 BM}, \fF_{B,1}, M\right) \right)+1\right]\notag\\&+\sum_{k=1}^d\frac{5136 d^4B^4}{M}\left[\log\left(14 \fN\left(\frac{ 1}{80 dBM}, \partial_k\fF_{B,1}, M\right) \right)+1\right]+3\inf_{\phi\in\fF_{B,1}}\|\phi-f\|_{W^{1,\infty}\left([0,1]^d\right)},
    \end{align}where\begin{align}\partial_k\fF_{B,1}&:=\{\bar{\phi}: \bar{\phi}=\partial_k\phi,\phi\in\fF_{B,1}\},\end{align}$\rmE$ is expected responding to $X$, and $X:=\{\vx_1,\ldots,\vx_M\}$ is an independent random variables set uniformly distributed on $[0,1]^d$.
			\end{proposition}

   \begin{proof}Due to the definition, we know that \[\fR_{D,1}(\vtheta_{D,1})=\inf_{\phi\in\fF_{B,1}}\|\phi-f\|_{H^{1}\left([0,1]^d\right)}.\] For the sample error, due to $\vtheta_{S,1}$ and $\vtheta_{D,1}$ belong to $\Theta$ almost surely, we have\begin{align}
					&  \rmE\left[\fR_{D,1}(\vtheta_{S,1})-\fR_{S,1}(\vtheta_{S,1})\right]\notag\\\le &\rmE\left[\int_{[0,1]^d}\left|\nabla f(\vx)-\nabla\phi(\vx;\vtheta_{S,1})\right|^2\,\D \vx -\frac{1}{M}\sum_{i=1}^M\left| \nabla f(\vx_i)-\nabla\phi(\vx_i;\vtheta_{S,1})\right|^2\right]+\rmE\left[\fR_{D,0}(\vtheta_{S,0})-\fR_{S,0}(\vtheta_{S,0})\right].
				\end{align} For the upper bound of \(\rmE\left[\fR_{D,0}(\vtheta_{S,0})-\fR_{S,0}(\vtheta_{S,0})\right]\) can be obtained by the Proposition \ref{connect rad l2}.

    For the rest of sample error, we have \begin{align}
        &\rmE\left[\int_{[0,1]^d}\left|\nabla f(\vx)-\nabla\phi(\vx;\vtheta_{S,1})\right|^2\,\D \vx -\frac{1}{M}\sum_{i=1}^M\left| \nabla f(\vx_i)-\nabla\phi(\vx_i;\vtheta_{S,1})\right|^2\right]\notag\\=&\rmE\left[\int_{[0,1]^d}\left|\nabla f(\vx)-\nabla\phi(\vx;\vtheta_{S,1})\right|^2\,\D \vx -\frac{2}{M}\sum_{i=1}^M\left| \nabla f(\vx_i)-\nabla\phi(\vx_i;\vtheta_{S,1})\right|^2\right]+\rmE\frac{1}{M}\sum_{i=1}^M\left| \nabla f(\vx_i)-\nabla\phi(\vx_i;\vtheta_{S,1})\right|^2\notag\\\le& \rmE\left[\int_{[0,1]^d}\left|\nabla f(\vx)-\nabla\phi(\vx;\vtheta_{S,1})\right|^2\,\D \vx -\frac{2}{M}\sum_{i=1}^M\left| \nabla f(\vx_i)-\nabla\phi(\vx_i;\vtheta_{S,1})\right|^2\right]+\inf_{\phi\in\fF_{B,1}}\|\phi-f\|_{W^{1,\infty}\left([0,1]^d\right)}
    \end{align}where the last inequality is due to the definition of $\vtheta_{S,1}$.

    Set \begin{align}
        \rmA_1&=\int_{[0,1]^d}\left|\nabla f(\vx)-\nabla\phi(\vx;\vtheta_{S,1})\right|^2\,\D \vx -\frac{2}{M}\sum_{i=1}^M\left| \nabla f(\vx_i)-\nabla\phi(\vx_i;\vtheta_{S,1})\right|^2\notag\\&=\sum_{k=1}^d\rmA_{1,k}\notag\\&:=\sum_{k=1}^d\left[\int_{[0,1]^d}\left|\partial_k f(\vx)-\partial_k\phi(\vx;\vtheta_{S,1})\right|^2\,\D \vx -\frac{2}{M}\sum_{i=1}^M\left| \partial_k f(\vx_i)-\partial_k\phi(\vx_i;\vtheta_{S,1})\right|^2\right].\notag
    \end{align}

    Note that \(\fR_{S,0}\left(\partial_k f\right)=0\) and\begin{align}
    \rmP(\rmA_{1,k}\ge \epsilon)&=\rmP\left( 2\|\partial_k f-\partial_k\phi\|_{L^2([0,1]^d)} -\frac{2}{M}\sum_{i=1}^M\left| \partial_k f(\vx_i)-\partial_k\phi(\vx_i;\vtheta_{S,1})\right|^2\ge\epsilon+\|\partial_k f-\partial_k\phi\|_{L^2([0,1]^d)}\right)\notag\\&=\rmP\left( 2\left(\|\partial_k f-\partial_k\phi\|_{L^2([0,1]^d)}-\left(\fR_{S,0}(\partial_k\phi)-\fR_{S,0}\left(\partial_k  f\right)\right)\right)\ge\epsilon+\|\partial_k f-\partial_k\phi\|_{L^2([0,1]^d)}\right)\notag\\&=\rmP\left( \|\partial_k f-\partial_k\phi\|_{L^2([0,1]^d)}-\left(\fR_{S,0}(\partial_k\phi)-\fR_{S,0}\left(\partial_k f\right)\right)\ge\frac{1}{2}\left(\frac{1}{2}\epsilon+\frac{1}{2}\epsilon+\|\partial_k f-\partial_k\phi\|_{L^2([0,1]^d)}\right)\right)\notag\\&\le 14 \fN\left(\frac{ \epsilon}{80 dB}, \partial_k\fF_{B,1}, M\right) \exp \left(-\frac{\epsilon M}{5136 d^4B^4}\right),
\end{align}where the last inequality is due to Lemma \ref{connect}.

Therefore, we have that \begin{align}
    \rmE \rmA_{1,k}&\le \int_{0}^\infty \rmP(\rmA_{1,k}\ge y)\,\D y\le\epsilon+\int_{\epsilon}^\infty \rmP(\rmA_{1,k}\ge y)\,\D y\notag\\&\le\epsilon+\int_{\epsilon}^\infty 14 \fN\left(\frac{ \epsilon}{80 dB}, \partial_k\fF_{B,1}, M\right) \exp \left(-\frac{y M}{5136 d^4B^4}\right)\,\D y
\end{align}

By the direct calculation, we have \begin{align}
    \int_{\epsilon}^\infty 14 \fN\left(\frac{ \epsilon}{80 dB}, \partial_k\fF_{B,1}, M\right) \exp \left(-\frac{y M}{5136 d^4B^4}\right)\,\D y\le 14 \fN\left(\frac{ \epsilon}{80 dB}, \partial_k\fF_{B,1}, M\right) \frac{5136 d^4B^4}{M}\exp \left(-\frac{\epsilon M}{5136 d^4B^4}\right).
\end{align} Set \[\epsilon=\frac{5136 d^4B^4}{M}\log\left(14 \fN\left(\frac{ 1}{80 dBM}, \partial_k\fF_{B,1}, M\right) \right)\ge \frac{1}{M}\] and we have \[\rmE \rmA_{1,k}\le \frac{5136 B^4d^4}{M}\left[\log\left(14 \fN\left(\frac{ 1}{80 BM}, \partial_k\fF_{B,1}, M\right) \right)+1\right]\]

Hence we have \begin{align}
        &\rmE\left[\fR_{D,1}(\vtheta_{S,1})-\fR_{S,1}(\vtheta_{S,1})\right]\le \frac{5136 B^4}{M}\left[\log\left(14 \fN\left(\frac{ 1}{80 BM}, \fF_{B,1}, M\right) \right)+1\right]\notag\\&+\sum_{k=1}^d\frac{5136 d^4B^4}{M}\left[\log\left(14 \fN\left(\frac{ 1}{80 dBM}, \partial_k\fF_{B,1}, M\right) \right)+1\right]+2\inf_{\phi\in\fF_{B,1}}\|\phi-f\|_{W^{1,\infty}\left([0,1]^d\right)}.
    \end{align}
    Finally, we set \(N = \fO(\log L)\) according to the definition of DeNNs, and the number of parameters \(W = \fO(N^2L\log L) = \fO(L(\log L)^3)\) to complete the proof.
			\end{proof}

   The remainder of the proof focuses on estimating $\fN\left(\frac{1}{80 dBM}, \partial_k\fF_{B,1}, M\right)$ with the aim of limiting the generalization error. The pseudo-dimension serves as a valuable tool for establishing such constraints. Notably, \cite{yang2023nearly} provides nearly optimal bounds for the pseudo-dimension in the context of DeNN derivatives. Drawing inspiration from the proof presented in \cite{yang2023nearly}, the upcoming propositions will illustrate the pseudo-dimension of $\partial_k\fF_{B,1}$.

   Before we estimate pseudo-dimension of $\partial_k\fF_{B,1}$, we first introduce Vapnik--Chervonenkis dimension (VC-dimension) \begin{definition}[VC-dimension \cite{abu1989vapnik}]
		Let $H$ denote a class of functions from $\fX$ to $\{0,1\}$. For any non-negative integer $m$, define the growth function of $H$ as \[\Pi_H(m):=\max_{x_1,x_2,\ldots,x_m\in \fX}\left|\{\left(h(x_1),h(x_2),\ldots,h(x_m)\right): h\in H \}\right|.\] The Vapnik--Chervonenkis dimension (VC-dimension) of $H$, denoted by $\text{VCdim}(H)$, is the largest $m$ such that $\Pi_H(m)=2^m$. For a class $\fG$ of real-valued functions, define $\text{VCdim}(\fG):=\text{VCdim}(\sgn(\fG))$, where $\sgn(\fG):=\{\sgn(f):f\in\fG\}$ and $\sgn(x)=1[x>0]$.\end{definition}

    In the proof of Proposition \ref{pdimh11}, we use the following lemmas:
		\begin{lemma}[{{\cite{bartlett2019nearly,anthony1999neural}}}]\label{bounded}
			Suppose $W\le M$ and let $P_1,\ldots,P_M$ be polynomials of degree at most $D$ in $W$ variables. Define \[K:=\left|\{\left(\sgn(P_1(a)),\ldots,\sgn(P_M(a))\right):a\in\sR^W\}\right|,\] then we have $K\le 2(2eMD/W)^W$.
		\end{lemma}
		
		\begin{lemma}[{\cite{bartlett2019nearly}}]\label{inequality}
			Suppose that $2^m\le 2^t(mr/w)^w$ for some $r\ge 16$ and $m\ge w\ge t\ge0$. Then, $m\le t+w\log_2(2r\log_2r)$.
		\end{lemma}

    \begin{proposition}
        \label{pdimh11}
			For any $N,L,d\in\sN_+$, there exists a constant $\bar{C}$ independent with $N,L$  such that 	\begin{equation}
				\text{Pdim}(\partial_k\fF_{B,1})\le \bar{C} N^2L^2\log_2 L\log_2 N
			\end{equation}for any $k=1,2,\ldots,d$.
		\end{proposition}
  \begin{proof}
  Denote \[\fF_{B,1,\fN}:=\{\eta(\vx,y):\eta(\vx,y)=\psi(\vx)-y,\psi\in \partial_k\fF_{B,1}, (\vx,y)\in\sR^{d+1}\}.\]
			Based on the definition of VC-dimension and pseudo-dimension, we have that\begin{equation}
				\text{Pdim}(\partial_k\fF_{B,1})\le \text{VCdim}(\fF_{B,1,\fN}).
			\end{equation}
			For the $\text{VCdim}(\fF_{B,1,\fN})$, it can be bounded by following way. The proof is similar to that in \cite{yang2023nearly,yang2023nearlys}.
   
			For a DeNN with $N$ width and $L$ depth, it can be represented as \[\phi=\vW_{L+1}\sigma_1(\vW_{L}\sigma_1(\ldots\sigma_1(\vW_1\vx+\vb_1)\ldots)+\vb_L)+b_{L+1}.\] Therefore, \begin{align}
				\psi(\vx)=\partial_k \phi(\vx)=&\vW_{L+1}\sigma_0(\vW_{L}\sigma_1(\ldots\sigma_1(\vW_1\vx+\vb_1)\ldots)+\vb_L)\notag\\&\cdot \vW_{L}\sigma_0(\ldots\sigma_1(\vW_1\vx+\vb_1)\ldots)\ldots\vW_2\sigma_0(\vW_1\vx+\vb_1)(\vW_1)_k,
			\end{align}where $\vW_i\in\sR^{N_i\times N_{i-1}}$ ($(\vW)_i$ is $i$-th column of $\vW$) and $\vb_i\in\sR^{N_i}$ are the weight matrix and the bias vector in the $i$-th linear transform in $\phi$, and $\sigma_0(x)=\sgn(x)=1[x>0],$ which is the derivative of the ReLU function and $~\sigma_0(\vx)=\diag(\sigma_0(x_i))$. 
			
			Let $\bar{\vx}=(\vx,y)\in\sR^{d+1}$ be an input and $\vtheta\in\sR^W$ be a parameter vector in $\eta:=\psi^2-y$. We denote the output of $\psi$ with input $\vx$ and parameter vector $\vtheta$ as $f(\vx,\vtheta)$. For fixed $\vx_1,\vx_2,\ldots,\vx_m$ in $\sR^d$, we aim to bound\begin{align}
				K:=\left|\{\left(\sgn(f(\vx_1,\vtheta)),\ldots,\sgn(f(\vx_m,\vtheta))\right):\vtheta\in\sR^W\}\right|.
			\end{align}
			
			The proof is inspired by \citep[Theorem 1]{yang2023nearly}. For any partition $\fS=\{P_1,P_2,\ldots,P_T\}$ of the parameter domain $\sR^W$, we have $K\le \sum_{i=1}^T\left|\{\left(\sgn(f(\vx_1,\vtheta)),\ldots,\sgn(f(\vx_m,\vtheta))\right):\vtheta\in P_i\}\right|$. We choose the partition such that within each region $P_i$, the functions $f(\vx_j,\cdot)$ are all fixed polynomials of bounded degree. This allows us to bound each term in the sum using Lemma \ref{bounded}.
			
			We define a sequence of sets of functions $\{\sF_j\}_{j=0}^L$ with respect to parameters $\vtheta\in\sR^W$:\begin{align}
				\sF_0&:=\{\vW_1\vx+\vb_1\}\cup \{(\vW_1)_k\}\notag\\
				\sF_1&:=\{\vW_2\sigma_0(\vW_1\vx+\vb_1),\vW_2\sigma_1(\vW_1\vx+\vb_1)+\vb_2\}\cup \{(\vW_1)_k\}\notag\\\sF_2&:=\{\vW_2\sigma_0(\vW_1\vx+\vb_1),\vW_3\sigma_0(\vW_2\sigma_1(\vW_1\vx+\vb_1)+\vb_2),\vW_3\sigma_1(\vW_2\sigma_1(\vW_1\vx+\vb_1)+\vb_2)+\vb_3\}\cup \{(\vW_1)_k\}\notag\\&\vdots\notag\\\sF_L&:=\{\vW_2\sigma_0(\vW_1\vx+\vb_1),\ldots,\vW_{L+1}\sigma_0(\vW_{L}\sigma_1(\ldots\sigma_1(\vW_1\vx+\vb_1)\ldots)+\vb_L)\}\cup \{(\vW_1)_k\}.
			\end{align}
			
			The partition of $\sR^W$ is constructed layer by layer through successive refinements denoted by $\fS_0,\fS_1,\ldots,\fS_L$. These refinements possess the following properties:

            \textbf{1}. We have $|\fS_0|=1$, and for each $n=1,\ldots,L$, we have $\frac{|\fS_n|}{|\fS_{n-1}|}\le 2\left(\frac{2emnN_k}{\sum_{i=1}^nW_i}\right)^{\sum_{i=1}^nW_i}$.
			
			\textbf{2}. For each $n=0,\ldots,L-1$, each element $S$ of $\fS_{n}$, when $\vtheta$ varies in $S$, the output of each term in $\sF_n$ is a fixed polynomial function in $\sum_{i=1}^nW_i$ variables of $\vtheta$, with a total degree no more than $n+1$.

            \textbf{3}. For each element $S$ of $\fS_{L}$, when $\vtheta$ varies in $S$, the $h$-th term in $\sF_L$ for $h\in\{1,2,\ldots,L+1\}$ is a fixed polynomial function in $W_{h}$ variables of $\vtheta$, with a total degree no more than $1$.
			
			We define $\fS_0=\{\sR^W\}$, which satisfies properties 1,2 above, since $\vW_1\vx_j+\vb_1$ and $(\vW_1)_i$ are affine functions of $\vW_1,\vb_1$.
			
			To define $\fS_n$, we use the last term of $\sF_{n-1}$ as inputs for the last two terms in $\sF_n$. Assuming that $\fS_0,\fS_1,\ldots,\fS_{n-1}$ have already been defined, we observe that the last two terms are new additions to $\sF_n$ when comparing it to $\sF_{n-1}$. Therefore, all elements in $\sF_n$ except the last two are fixed polynomial functions in $W_n$ variables of $\vtheta$, with a total degree no greater than $n$ when $\vtheta$ varies in $S\in\fS_n$. This is because $\fS_n$ is a finer partition than $\fS_{n-1}$.
			
			We denote $p_{\vx_j,n-1,S,k}(\vtheta)$ as the output of the $k$-th node in the last term of $\sF_{n-1}$ in response to $\vx_j$ when $\vtheta\in S$. The collection of polynomials \[\{p_{\vx_j,n-1,S,k}(\vtheta): j=1,\ldots,m,~k=1,\ldots,N_n\}\]can attain at most $2\left(\frac{2emnN_n}{\sum_{i=1}^nW_i}\right)^{\sum_{i=1}^nW_i}$ distinct sign patterns when $\vtheta\in S$ due to Lemma \ref{bounded} for sufficiently large $m$. Therefore, we can divide $S$ into $2\left(\frac{2emnN_n}{\sum_{i=1}^nW_i}\right)^{\sum_{i=1}^nW_i}$ parts, each having the property that $p_{\vx_j,n-1,S,k}(\vtheta)$ does not change sign within the subregion. By performing this for all $S\in\fS_{n-1}$, we obtain the desired partition $\fS_n$. This division ensures that the required property 1 is satisfied.
			
			Additionally, since the input to the last two terms in $\sF_n$ is $p_{\vx_j,n-1,S,k}(\vtheta)$, and we have shown that the sign of this input will not change in each region of $\fS_n$, it follows that the output of the last two terms in $\sF_n$ is also a polynomial without breakpoints in each element of $\fS_n$. Therefore, the required property 2 is satisfied.

            In the context of DeNNs, the last layer is characterized by all terms containing the activation function $\sigma_0$. Consequently, for any element $S$ of the partition $\fS_{L}$, when the vector of parameters $\vtheta$ varies within $S$, the $h$-th term in $\sF_L$ for $h\in\{1,2,\ldots,L+1\}$ can be expressed as a polynomial function of at most degree $1$, which depends on at most $W_{h}$ variables of $\vtheta$. Hence, the required property 3 is satisfied.
			
			Due to property 3, note that there is a partition $\fS$ where \[|\fS|\le \prod_{n=1}^{L}2\left(\frac{2emnN_n}{\sum_{i=1}^nW_i}\right)^{\sum_{i=1}^nW_i}.\]And the output of $\psi$ is a polynomial function in $\sum_{i=1}^{L+1}W_i$ variables of $\vtheta\in S\in\fS_L$, of total degree no more than $L+1$. Therefore, for each $S\in\fS_L$ we have \[\left|\{\left(\sgn(f(\vx_1,\vtheta)),\ldots,\sgn(f(\vx_m,\vtheta))\right):\vtheta\in S\}\right|\le 2\left(2em(L+1)/\sum_{i=1}^{L+1}W_i\right)^{\sum_{i=1}^{L+1}W_i}.\] Then \begin{align}
				K\le 2^{L+1} \left(\frac{em(L+1)(L+2)N}{U}\right)^{U}
			\end{align}where $U:=\sum_{n=1}^{L}\sum_{i=1}^nW_i=\fO(N^2L^2)$, $N$ is the width of the network, and the last inequality is due to weighted AM-GM. For the definition of the VC-dimension, we have \begin{equation}
				2^{\text{VCdim}(D\fF_{B,1,\fN})}\le 2^{L} \left(\frac{e\text{VCdim}(D\fF_{B,1,\fN})(L+1)LN}{U}\right)^{U}.
			\end{equation}Due to Lemma \ref{inequality}, we obtain that\begin{equation}
				\text{VCdim}(\fF_{B,1,\fN})\le L+1+U\log_2[2(L+1)(L+2)\log_2(L+1)(L+2) ]=\fO(N^2L^2\log_2 L\log_2 N)
			\end{equation}since $U=\fO(N^2L^2)$.
		\end{proof}

   \begin{proof}[Proof of Theorem \ref{generalzation l2} for $k=1$]
       The proofs of Theorem \ref{generalzation l2} for $k=1$ are analogous to those of Theorem \ref{generalzation l2} for $k=0$ by combining the results of Propositions \ref{connect radh1} and \ref{pdimh11}.
   \end{proof}

   \subsection{Proof of Theorem \ref{generalzation l2} for $k=2$}
The Theorem \ref{generalzation l2} in the \(k=2\) case can be read as follows:


   \begin{proposition}
			\label{connect radh2}
				Let $d, N, L,M\in\sN_+$, $B,C_1,C_2\in\sR_+$. For any $f\in W^{n,\infty}([0,1]^d)$ with $\|f\|_{W^{n,\infty}([0,1]^d)}\le 1$, we have \begin{align}
        &\rmE\fR_{D,2}(\vtheta_{S,2})\le \frac{5136 B^4}{M}\left[\log\left(14 \fN\left(\frac{ 1}{80 BM}, \fF_{B,2}, M\right) \right)+1\right]\notag\\&+\frac{5136 d^4B^4}{M}\left[\log\left(14 \fN\left(\frac{ 1}{80 dBM}, \Delta\fF_{B,2}, M\right) \right)+1\right]+3\inf_{\phi\in\fF_{B,2}}\|\phi-f\|_{W^{2,\infty}\left([0,1]^d\right)},
    \end{align}where\begin{align}\Delta\fF_{B,2}&:=\{\bar{\phi}: \bar{\phi}=\Delta\phi,\phi\in\fF_{B,2}\},\end{align}$\rmE$ is expected responding to $X$, and $X:=\{\vx_1,\ldots,\vx_M\}$ is an independent random variables set uniformly distributed on $[0,1]^d$.
			\end{proposition}

   \begin{proof}The proof closely resembles that of Proposition \ref{connect radh1}.
			\end{proof}

 The subsequent part of the proof centers on estimating $\fN\left(\frac{1}{80 dBM}, \Delta\fF_{B,2}, M\right)$ to constrain the generalization error. Utilizing the pseudo-dimension proves instrumental in establishing these bounds.

    \begin{proposition}\label{pdimh21}
			For any $N,L,d\in\sN_+$, there exists a constant $\bar{C}$ independent with $N,L$  such that 	\begin{equation}
				\text{Pdim}(\Delta\fF_{B,2})\le \bar{C} N^2L^2\log_2 L\log_2 N.
			\end{equation}
		\end{proposition}	
  \begin{proof}
  Denote \[\fF_{B,2,\fN}:=\{\eta(\vx,y):\eta(\vx,y)=\psi(\vx)-y,\psi\in \Delta\fF_{B,2}, (\vx,y)\in\sR^{d+1}\}.\]
			Based on the definition of VC-dimension and pseudo-dimension, we have that\begin{equation}
				\text{Pdim}(\Delta\fF_{B,2})\le \text{VCdim}(\fF_{B,2,\fN}).
			\end{equation}
			For the $\text{VCdim}(\fF_{B,2,\fN})$, it can be bounded by following way. The proof is similar to that in \cite{yang2023nearlys}.
   
			For a DeNN with $N$ width and $L$ depth, it can be represented as \[\phi=\vW_{L+1}\sigma(\vW_{L}\sigma(\ldots\sigma(\vW_1\vx+\vb_1)\ldots)+\vb_L)+b_{L+1},\]where $\sigma$ can be either the ReLU or the ReLU square. Then the first order derivative can be read as \begin{align}
	\psi(\vx)=D_i \phi(\vx)=&\vW_{L+1}\sigma'(\vW_{L}\sigma(\ldots\sigma(\vW_1\vx+\vb_1)\ldots)+\vb_L)\notag\\&\cdot \vW_{L}\sigma'(\ldots\sigma(\vW_1\vx+\vb_1)\ldots)\ldots\vW_2\sigma'(\vW_1\vx+\vb_1)(\vW_1)_i,
\end{align}where $\vW_i\in\sR^{N_i\times N_{i-1}}$ ($(\vW)_i$ is $i$-th column of $\vW$) and $\vb_i\in\sR^{N_i}$ are the weight matrix and the bias vector in the $i$-th linear transform in $\phi$, and $\sigma'(\vx)=\diag(\sigma'(x_i))$. Then we have \begin{align}
\lambda(\vx)=\Delta\phi(\vx)=\sum_{a=1}^{L+1} \lambda_a(\vx).
\end{align} For \begin{align}
\lambda_a(\vx)=&\sum_{i=1}^d\vW_{L+1}\sigma'(\vW_{L}\sigma(\ldots\sigma(\vW_1\vx+\vb_1)\ldots)+\vb_L)\cdot  \vW_{L}\sigma'(\ldots\sigma(\vW_1\vx+\vb_1)\ldots)\ldots\notag\\&\cdot[\vW_{a}\sigma''(\ldots\sigma(\vW_1\vx+\vb_1)\ldots)\cdot \vW_{a-1}\sigma'(\ldots\sigma(\vW_1\vx+\vb_1)\ldots)\vW_2\sigma'(\vW_1\vx+\vb_1)(\vW_1)_i]\notag\\&\cdot\ldots\vW_{a-1}\sigma'(\ldots\sigma(\vW_1\vx+\vb_1)\ldots)\vW_2\sigma'(\vW_1\vx+\vb_1)(\vW_1)_i,\label{gamma}
\end{align}where $\sigma''(\vx)=\diag(\sigma''(x_i))$ is a three-order tensor. Denote $W_i$ as the number of parameters in $\vW_i,\vb_i$, i.e., $W_i=N_iN_{i-1}+N_i$.
			
			Let $\bar{\vx}=(\vx,y)\in\sR^{d+1}$ be an input and $\vtheta\in\sR^W$ be a parameter vector in $\eta:=\lambda^2-y$. We denote the output of $\lambda_a$ with input $\vx$ and parameter vector $\vtheta$ as $f(\vx,\vtheta)$. For fixed $\vx_1,\vx_2,\ldots,\vx_m$ in $\sR^d$, we aim to bound\begin{align}
				K:=\left|\{\left(\sgn(f(\vx_1,\vtheta)),\ldots,\sgn(f(\vx_m,\vtheta))\right):\vtheta\in\sR^W\}\right|.
			\end{align}
			
			The proof is inspired by \citep[Theorem 2]{yang2023nearlys}. For any partition $\fS=\{P_1,P_2,\ldots,P_T\}$ of the parameter domain $\sR^W$, we have \[K\le \sum_{i=1}^T\left|\{\left(\sgn(f(\vx_1,\vtheta)),\ldots,\sgn(f(\vx_m,\vtheta))\right):\vtheta\in P_i\}\right|.\] We choose the partition such that within each region $P_i$, the functions $f(\vx_j,\cdot)$ are all fixed polynomials of bounded degree. This allows us to bound each term in the sum using Lemma \ref{bounded}.

  We define a sequence of sets of functions $\{\sF_j\}_{j=0}^L$ with respect to parameters $\vtheta\in\sR^W$:\begin{align}
				\sF_0:=\{&(\vW_1)_1,(\vW_1)_2,\ldots,(\vW_1)_d,\vW_1\vx+\vb_1\}\notag\\
				\sF_1:=\{&\vW_2\sigma''(\vW_1\vx+\vb_1),\vW_2\sigma'(\vW_1\vx+\vb_1),\vW_2\sigma(\vW_1\vx+\vb_1)+\vb_2\}\cup \sF_0\notag\\\sF_2:=\{&\vW_3\sigma''(\vW_2\sigma(\vW_1\vx+\vb_1)+\vb_2),\vW_3\sigma'(\vW_2\sigma(\vW_1\vx+\vb_1)+\vb_2),\notag\\&\vW_3\sigma(\vW_2\sigma(\vW_1\vx+\vb_1)+\vb_2)+\vb_3\}\cup \sF_1\notag\\&\vdots\notag\\\sF_L:=\{&\vW_{L+1}\sigma'(\vW_{L}\sigma_1(\ldots\sigma_1(\vW_1\vx+\vb_1)\ldots)+\vb_L),\notag\\&\vW_{L+1}\sigma''(\vW_{L}\sigma_1(\ldots\sigma_1(\vW_1\vx+\vb_1)\ldots)+\vb_L)\}\cup \sF_{L-1}.
			\end{align}

   The partition of $\sR^W$ is constructed layer by layer through successive refinements denoted by $\fS_0,\fS_1,\ldots,\fS_L$. We denote $L^*=L-C\log_2 L$. These refinements possess the following properties:

            \textbf{1}. We have $|\fS_0|=1$, and for each $n=1,\ldots,L$, we have \[\frac{|\fS_n|}{|\fS_{n-1}|}\le 2\left(\frac{2em(1+(n-1)2^{\max\{0,n-1-L^*\}})N_n}{\sum_{i=1}^nW_i}\right)^{\sum_{i=1}^nW_i}.\]
			
			\textbf{2}. For each $n=0,\ldots,L^*$, each element $S$ of $\fS_{n}$, when $\vtheta$ varies in $S$, the output of each term in $\sF_n$ is a fixed polynomial function in $\sum_{i=1}^nW_i$ variables of $\vtheta$, with a total degree no more than $1+n2^{\max\{0,n-L^*\}}$.
			
			We define $\fS_0=\{\sR^W\}$, which satisfies properties 1,2 above, since $\vW_1\vx_j+\vb_1$ and $(\vW_1)_i$ for all $i=1,\ldots,d$ are affine functions of $\vW_1,\vb_1$.
			
			For each $n=0,\ldots,L$, to define $\fS_n$, we use the last term of $\sF_{n-1}$ as inputs for the new terms in $\sF_n$. Assuming that $\fS_0,\fS_1,\ldots,\fS_{n-1}$ have already been defined, we observe that the last two or three terms are new additions to $\sF_n$ when comparing it to $\sF_{n-1}$. Therefore, all elements in $\sF_n$ except the $\sF_n\backslash \sF_{n-1}$ are fixed polynomial functions in $W_n$ variables of $\vtheta$, with a total degree no greater than $1+(n-1)2^{\max\{0,n-1-L^*\}}$ when $\vtheta$ varies in $S\in\fS_n$. This is because $\fS_n$ is a finer partition than $\fS_{n-1}$.
			
			We denote $p_{\vx_j,n-1,S,k}(\vtheta)$ as the output of the $k$-th node in the last term of $\sF_{n-1}$ in response to $\vx_j$ when $\vtheta\in S$. The collection of polynomials \[\{p_{\vx_j,n-1,S,k}(\vtheta): j=1,\ldots,m,~k=1,\ldots,N_n\}\]can attain at most $2\left(\frac{2em(1+(n-1)2^{\max\{0,n-1-L^*\}})N_n}{\sum_{i=1}^nW_i}\right)^{\sum_{i=1}^nW_i}$ distinct sign patterns when $\vtheta\in S$ due to Lemma \ref{bounded} for sufficiently large $m$. Therefore, we can divide $S$ into \[2\left(\frac{2em(1+(n-1)2^{\max\{0,n-1-L^*\}})N_n}{\sum_{i=1}^nW_i}\right)^{\sum_{i=1}^nW_i}\] parts, each having the property that $p_{\vx_j,n-1,S,k}(\vtheta)$ does not change sign within the subregion. By performing this for all $S\in\fS_{n-1}$, we obtain the desired partition $\fS_n$. This division ensures that the required property 1 is satisfied.
			
			Additionally, since the input to the last two terms in $\sF_n$ is $p_{\vx_j,n-1,S,k}(\vtheta)$, and we have shown that the sign of this input will not change in each region of $\fS_n$, it follows that the output of the last two terms in $\sF_n$ is also a polynomial without breakpoints in each element of $\fS_n$, therefore, the required property 2 is satisfied.

			
			Due to the structure of $\lambda$, note that there is a partition $\fS$ where \[|\fS|\le 2\left(\frac{2em(1+(n-1)2^{\max\{0,n-1-L^*\}})N_n}{\sum_{i=1}^nW_i}\right)^{\sum_{i=1}^nW_i},\]where $L^*=10(L+1)\log_24L-6\log_2(10(L+1)\log_24L)$. And the output of $\lambda$ is a polynomial function in $\sum_{i=1}^{L+1}W_i$ variables of $\vtheta\in S\in\fS_L$, of total degree no more than \[d_2:=2\sum_{n=0}^L(1+n2^{\max\{0,n-L^*\}})=2L+2+L^6(L-1).\] The rest of the proof is similar with those in Proposition \ref{pdimh11}, we can obtain that\begin{equation}
				\text{VCdim}(\fF_{B,2,\fN})\le \fO(N^2L^2\log_2 L\log_2 N).
			\end{equation}
		\end{proof}

  \begin{proof}[Proof of Theorem \ref{generalzation l2} for $k=2$]
       The proofs of Theorem \ref{generalzation l2} for $k=2$ are analogous to those of Theorem \ref{generalzation l2} for $k=0$ by combining the results of Propositions \ref{connect radh2} and \ref{pdimh21}.
   \end{proof}

   \section{Proofs of Theorems \ref{generalzation r} and \ref{generalzation p}}

  The sample errors in Theorems \ref{generalzation r} and \ref{generalzation p} can be obtained by applying Theorem\ref{generalzation l2}. The remaining problem is to establish the approximation error. For the approximations in Theorems \ref{generalzation r}, it can be estimated by the Lemma \ref{app r}.

   \begin{lemma}
    Suppose $f\in L^2(\widehat{\Omega})$ and $\widehat{\Omega}$ denotes the domain belonging to $(0,1)^d$ with a smooth boundary, then $u^*(\vx)\in H^2(\widehat{\Omega})$ where $u^*(\vx)$ is the exact solution of Eq.~(\ref{PDE}).
\end{lemma}

\begin{proof}
    The proof can be found in \cite{evans2022partial}.
\end{proof}

For the approximation error in Theorem~\ref{generalzation p}, we can rewrite it by the following lemma (Lemma \ref{H2}):\begin{lemma}
     Suppose that $\phi(\vx;\vtheta_{D,p})\in \fF_{B,2}$ and $u^*$ is the exact solution of Eq.~(\ref{PDE}) with Dirichlet boundary condition, then there is a constant $C$ such that\[\fE_{D,p}(\phi(\vx;\vtheta_{D,p}))\le C\|\phi(\vx;\vtheta_{D,p})-u^*\|^2_{H^2(\widehat{\Omega})}.\]
\end{lemma} 
\begin{proof}
    Due to $u^*$ is the exact solution and trace theorem inequality, we know that \begin{align}
\fE_{D,p}(\phi) &:=  \int_{\widehat{\Omega}} |\Delta \phi(\vx;\vtheta)-\Delta u^*(\vx)|^2 \, \D \vx + \lambda \int_{\partial \widehat{\Omega}} |\phi(\vx;\vtheta)-u^*(\vx)|^2 \, \D s,\notag\\&\le C\|\phi(\vx;\vtheta)-u^*(\vx)\|_{H^2(\widehat{\Omega})}^2
\end{align}
\end{proof}

\section{Proof of Theorem \ref{generalzation l2R}}
\subsection{Rademacher complexity and related lemmas}

  \begin{definition}
  [Rademacher complexity \cite{anthony1999neural}]
				Given a sample set $S=\{z_1,z_2,\ldots,z_M\}$ on a domain $\fZ$, and a class $\fF$ of real-valued functions defined on $\fZ$, the empirical Rademacher complexity of $\fF$ in $S$ is defined as \[\tR_S(\fF):=\frac{1}{M}\rmE_{\Sigma_M}\left[\sup_{f\in\fF}\sum_{i=1}^M\sigma_if(z_i)\right],\]where $\Sigma_M:=\{\sigma_1,\sigma_2,\ldots,\sigma_M\}$ are independent random variables drawn from the Rademacher distribution, i.e., $\rmP(\sigma_i=+1)=\rmP(\sigma_i=-1)=\frac{1}{2}$ for $i=1,2,\ldots,M.$ For simplicity, if $S=\{z_1,z_2,\ldots,z_M\}$ is an independent random variable set with the uniform distribution, denote \(\tR_M(\fF):=\rmE_S\tR_S(\fF).\)
			\end{definition}

   The following lemma will be used to bounded generalization error by Rademacher complexities:
			
			\begin{lemma}[\cite{wainwright2019high}, Proposition 4.11]\label{connect1}
				Let $\fF$ be a set of functions. Then \[\rmE_X\sup_{u\in\fF}\left|\frac{1}{M}\sum_{i=1}^Mu(x_j)-\rmE_{x\sim\fP_\Omega} u(x)\right|\le 2\tR_M(\fF),\]where $X:=\{x_1,\ldots,x_M\}$ is an independent random variable set with the uniform distribution.
			\end{lemma}
			
			 \begin{lemma}[{\cite{jiao2021error}}]\label{jiao}
			    Assume that $w: \Omega \rightarrow \sR$ and $|w(\vx)| \leq B$ for all $\vx \in \Omega$, then for any function
class $\fF$, there holds
$$
\tR_M(w \cdot \fF) \leq B\tR_M(\fF)
$$
where $w \cdot \fF:=\{\bar{u}: \bar{u}(\vx)=w(\vx) u(\vx), u \in \fF\}$.
			\end{lemma}

   Next, we present a lemma of uniform covering numbers. These concepts are essential for estimating the Rademacher complexity in the context of generalization error. Subsequently, we derive an estimate for the uniform covering number, leveraging the concept of pseudo-dimension.

			
			\begin{lemma}[Dudley's theorem \cite{anthony1999neural}]\label{dudley}
				Let $\fF$ be a function class such that $\sup_{f\in\fF}\|f\|_\infty\le B$. Then the Rademacher complexity $\tR_M(\fF)$ satisfies that \[\tR_M(\fF) \leq \inf _{0 \leq \delta \leq B}\left\{4 \delta+\frac{12}{\sqrt{M}} \int_\delta^B \sqrt{\log 2\fN(\varepsilon,\fF,M)} \,\D \varepsilon\right\}\]
			\end{lemma}

    \subsection{Proof of Theorem \ref{generalzation l2R}}
There are three cases in Theorem \ref{generalzation l2R}. In order to prove Theorem \ref{generalzation l2R}, we consider these three cases individually. For simplicity in notation, we still denote the width as \(N = \fO(\log L)\) in \(\mathcal{F}_{B,k}\) in the following proofs.

\subsubsection{Proof of Theorem \ref{generalzation l2R} for $k=0$.}
		
			 We first show the proof of the Proposition \ref{connect rad l2R}:
    			\begin{proposition}
			\label{connect rad l2R}
				Let $d, N, L,M\in\sN_+$, $B,C_1,C_2\footnote{Here, the occurrences of $C_1$ and $C_2$ in $\fB_{B,0}$ continue to hold throughout the rest of the propositions and theorems in this paper.}\in\sR_+$. For any $f\in W^{n,\infty}([0,1]^d)$ with $\|f\|_{W^{n,\infty}([0,1]^d)}\le 1$, we have \begin{align}&\rmE\fR_{D,0}(\vtheta_{S,0})\notag\\\le& \inf_{\phi\in\fF_{B,0}}\|\phi-f\|_{L^{2}\left([0,1]^d\right)}+4\tR_M(\fF_{B,0})+2\tR_M(\fF^2_{B,0}),\notag\end{align}where \(\fF^2_{B,0}:=\{\bar{\phi}: \bar{\phi}=\phi^2,\phi\in\fF_{B,0}\},\) $\rmE$ is expected responding to $X$, and $X:=\{\vx_1,\ldots,\vx_M\}$ is an independent random variables set uniformly distributed on $[0,1]^d$.
			\end{proposition}
\begin{proof}
 Due to the definition, we know that \[\fR_{D,0}(\vtheta_{D,0})=\inf_{\phi\in\fF_{B,0}}\|\phi-f\|_{L^{2}\left([0,1]^d\right)}.\] For the sample error, due to $\vtheta_{S,0}$ and $\vtheta_{D,0}$ belong to $\Theta$ almost surely, we have\begin{align}
					&  \left|\rmE\left[\fR_{S,0}(\vtheta_{S,0})-\fR_{D,0}(\vtheta_{S,0})\right]\right|\notag\\=&\left|\rmE\left[\frac{1}{M}\sum_{i=1}^M\left| f(\vx_i)-\phi(\vx_i;\vtheta_{S,0})\right|^2-\int_{[0,1]^d}\left| f(\vx)-\phi(\vx;\vtheta_{S,0})\right|^2\,\D \vx\right]\right|\notag\\\le&2\left|\rmE\left(\frac{1}{M}\sum_{i=1}^Mf(\vx_i)\phi(\vx_i;\vtheta_{S,0})-\int_{[0,1]^d} f(\vx)\phi(\vx;\vtheta_{S,0})\,\D \vx\right)\right|\notag\\&+ \left|\rmE\left(\frac{1}{M}\sum_{i=1}^M\phi(\vx_i;\vtheta_{S,0})^2-\int_{[0,1]^d} \phi(\vx;\vtheta_{S,0})^2\,\D \vx\right)\right|\notag\\\le&2\tR_M(f\fF_{B,0})+\tR_M(\fF^2_{B,0})
				\end{align}where the last inequality is due to Lemma \ref{connect1}. 

    Due to Lemmas \ref{connect1} and \ref{jiao}, and $|f|\le 1$, we have that \begin{align}
        \rmE\left|\frac{1}{M}\sum_{i=1}^Mf(\vx_i)\phi(\vx_i;\vtheta_{S,0})-\int_{[0,1]^d} f(\vx)\phi(\vx;\vtheta_{S,0})\,\D \vx)\right|\le 2\tR_M(f\fF_{B,0})\le 2\tR_M(\fF_{B,0}). 
    \end{align}

    Therefore, we have \begin{align}
        \rmE\fR_{S,0}(\vtheta_{S,0})-\fR_{D,0}(\vtheta_{S,0})\le 4\tR_M(\fF_{B,0})+2\tR_M(\fF^2_{B,0}).
    \end{align}
			\end{proof}

Next we need to bound $\text{Pdim}(\fF_{B,0})$ and $\text{Pdim}(\fF_{B,0}^2)$. Based on \cite{bartlett2019nearly}, $\text{Pdim}(\fF_{B,0})=\fO(L^2N^2\log_2 L\log_2 N)$. For the $\text{Pdim}(\fF_{B,0}^2)$, we can estimate it by the simlaler way of $\text{Pdim}(\fF_{B,0})$.

   \begin{proposition}\label{pdim}
			For any $N,L,d\in\sN_+$, there exists a constant $\widehat{C}$ independent with $N,L$  such that 	\begin{equation}
				\text{Pdim}(\fF_{B,0}^2)\le \widehat{C} N^2L^2\log_2 L\log_2 N.
			\end{equation}
		\end{proposition}
  
  \begin{proof}
  Denote \[\fF_{B,0,\fN}^2:=\{\eta(\vx,y):\eta(\vx,y)=\psi(\vx)-y,\psi\in \fF_{B,0}^2, (\vx,y)\in\sR^{d+1}\}.\]
			Based on the definition of VC-dimension and pseudo-dimension, we have that\begin{equation}
				\text{Pdim}(\fF_{B,0}^2)\le \text{VCdim}(\fF_{B,0,\fN}^2).
			\end{equation}
			For the $\text{VCdim}(\fF_{B,0,\fN}^2)$, it can be bounded by following way. The proof is similar to that in \cite{bartlett2019nearly}.
   
			For a DeNN with $N$ width and $L$ depth, it can be represented as \[\phi=\vW_{L+1}\sigma_1(\vW_{L}\sigma_1(\ldots\sigma_1(\vW_1\vx+\vb_1)\ldots)+\vb_L)+b_{L+1},\] where $\vW_i\in\sR^{N_i\times N_{i-1}}$ ($(\vW)_i$ is $i$-th column of $\vW$) and $\vb_i\in\sR^{N_i}$ are the weight matrix and the bias vector in the $i$-th linear transform in $\phi$. Denote $W_i$ as the number of parameters in $\vW_i,\vb_i$, i.e., $W_i=N_iN_{i-1}+N_i$. 
			
			Let $\bar{\vx}=(\vx,y)\in\sR^{d+1}$ be an input and $\vtheta\in\sR^W$ be a parameter vector in $\psi:=\phi^2-y$. We denote the output of $\psi$ with input $\vx$ and parameter vector $\vtheta$ as $f(\vx,\vtheta)$. For fixed $\vx_1,\vx_2,\ldots,\vx_m$ in $\sR^d$, we aim to bound\begin{align}
				K:=\left|\{\left(\sgn(f(\vx_1,\vtheta)),\ldots,\sgn(f(\vx_m,\vtheta))\right):\vtheta\in\sR^W\}\right|.
			\end{align}
			
			The proof is inspired by \citep[Theorem 7]{bartlett2019nearly}. For any partition $\fS=\{P_1,P_2,\ldots,P_T\}$ of the parameter domain $\sR^W$, we have $K\le \sum_{i=1}^T\left|\{\left(\sgn(f(\vx_1,\vtheta)),\ldots,\sgn(f(\vx_m,\vtheta))\right):\vtheta\in P_i\}\right|$. We choose the partition such that within each region $P_i$, the functions $f(\vx_j,\cdot)$ are all fixed polynomials of bounded degree. This allows us to bound each term in the sum using Lemma \ref{bounded}.
			
			Due to proof of \citep[Theorem 7]{bartlett2019nearly}, note that there is a partition $\fS$ where \[|\fS|\le \prod_{n=1}^{L-1}2\left(\frac{2emnN_n}{\sum_{i=1}^nW_i}\right)^{\sum_{i=1}^nW_i}.\]And the output $\phi$ is a polynomial function in $\sum_{i=1}^{L}W_i$ variables of $\vtheta\in S\in\fS_L$, of total degree no more than $L$. Hence the output $\psi$ is a polynomial function in $\sum_{i=1}^{L}W_i$ variables of $\vtheta\in S\in\fS_L$, of total degree no more than $2L$. Therefore, for each $S\in\fS_L$ we have \[\left|\{\left(\sgn(f(\vx_1,\vtheta)),\ldots,\sgn(f(\vx_m,\vtheta))\right):\vtheta\in S\}\right|\le 2\left(4emL/\sum_{i=1}^{L}W_i\right)^{\sum_{i=1}^{L}W_i}.\] Then \begin{align}
				K\le& 2\left(4emL/\sum_{i=1}^{L}W_i\right)^{\sum_{i=1}^{L}W_i}\cdot  \prod_{n=1}^{L}2\left(\frac{2emnN_n}{\sum_{i=1}^nW_i}\right)^{\sum_{i=1}^nW_i}\le \prod_{n=1}^{L}2\left(\frac{4emnN_n}{\sum_{i=1}^nW_i}\right)^{\sum_{i=1}^nW_i}\notag\\\le& 2^{L} \left(\frac{2em(L+1)LN}{U}\right)^{U}
			\end{align}where $U:=\sum_{n=1}^{L}\sum_{i=1}^nW_i=\fO(N^2L^2)$, $N$ is the width of the network, and the last inequality is due to weighted AM-GM. For the definition of the VC-dimension, we have \begin{equation}
				2^{\text{VCdim}(\fF_{B,0,\fN}^2)}\le 2^{L} \left(\frac{2e\text{VCdim}(\fF_{B,0,\fN}^2)(L+1)LN}{U}\right)^{U}.
			\end{equation}Due to Lemma \ref{inequality}, we obtain that\begin{equation}
				\text{VCdim}(\fF_{B,0,\fN}^2)\le L+U\log_2[4(L+1)L\log_2(L+1)L ]=\fO(N^2L^2\log_2 L\log_2 N)
			\end{equation}since $U=\fO(N^2L^2)$.
		\end{proof}

  Now we can show the proof of Theorem \ref{generalzation l2} for $k=0$:\begin{proof}[Proof of Theorem \ref{generalzation l2} for $k=0$]
				 Let $J=\max\{\text{Pdim}(\fF_{B,0}^2),\text{Pdim}(\fF_{B,0})\}$. Due to Lemma \ref{dudley}, \ref{cover dim} and Theorem \ref{pdim}, for any $M\ge J$, we have \begin{align}
					\tR_M(\fF_{B,0}^2)\le& 4\delta+\frac{12}{\sqrt{M}} \int_\delta^B \sqrt{\log 2\fN(\varepsilon,\fF_{B,0}^2,M)} \,\D \varepsilon\notag\\\le &4\delta+\frac{12}{\sqrt{M}} \int_\delta^B \sqrt{\log 2\left(\frac{2eMB}{\varepsilon\text{Pdim}(\fF_{B,0}^2)}\right)^{\text{Pdim}(\fF_{B,0}^2)}} \,\D \varepsilon\notag\\\le&4\delta+\frac{12B}{\sqrt{M}}+12\left(\frac{\text{Pdim}(\fF_{B,0}^2)}{M}\right)^{\frac{1}{2}} \int_\delta^B \sqrt{\log \left(\frac{2eMB}{\varepsilon\text{Pdim}(\fF_{B,0}^2)}\right)} \,\D \varepsilon.
				\end{align}
				
				By the direct calculation for the integral, we have\[\int_\delta^B \sqrt{\log \left(\frac{2eMB}{\varepsilon\text{Pdim}(\fF_{B,0}^2)}\right)} \,\D \varepsilon\le B\sqrt{\log \left(\frac{2eMB}{\delta\text{Pdim}(\fF_{B,0}^2)}\right)}.\]
				
				Then choosing $\delta=B\left(\frac{\text{Pdim}(\fF_{B,0}^2)}{M}\right)^{\frac{1}{2}}\le B$, we have \begin{equation}
					\tR_M(\fF_{B,0}^2)\le 28 B\left(\frac{\text{Pdim}(\fF_{B,0}^2)}{M}\right)^{\frac{1}{2}}\sqrt{\log \left(\frac{2eM}{\text{Pdim}(\fF_{B,0}^2)}\right)}.
				\end{equation}
				
				Therefore, due to Theorem \ref{pdim}, there is a constant $C_4$ independent with $L, N, M$ such as\begin{equation}
					\tR_M(\fF_{B,0}^2)\le C_4 \frac{NL(\log_2 L\log_2 N)^{\frac{1}{2}}}{\sqrt{M}}\log M.
				\end{equation}
				
				$\tR_M(\fF_{B,0})$ can be estimate in the similar way. Therefore, we have that there is a constant $C_5=C_5(B,d, C_1,C_2)$ such that\begin{equation}
					\rmE\fR_{D,0}(\vtheta_{S,0})\le C_{3,0} N^{-2 n / d} L^{-2 n / d}+C_5\frac{NL(\log_2 L\log_2 N)^{\frac{1}{2}}}{\sqrt{M}}\log M.
				\end{equation}
    Finally, we set \(N = \fO(\log L)\) according to the definition of DeNNs, and the number of parameters \(W = \fO(N^2L\log L) = \fO(L(\log L)^3)\) to complete the proof.
			\end{proof}

   \subsubsection{Proof of Theorem \ref{generalzation l2R} for $k=1$}

 \begin{proposition}
			\label{connect radh1R}
				Let $d, N, L,M\in\sN_+$, $B,C_1,C_2\in\sR_+$. For any $f\in W^{n,\infty}([0,1]^d)$ with $\|f\|_{W^{n,\infty}([0,1]^d)}\le 1$, we have \[\rmE\fR_{D,1}(\vtheta_{S,1})\le \inf_{\phi\in\fF_{B,1}}\|\phi-f\|_{H^1\left([0,1]^d\right)}+\fE_1+\fE_0,\]where \(\fE_1=4d\tR_M(D\fF_{B,1})+2\tR_M(D\fF^2_{B,1})\) and \(\fE_0=4\tR_M(\fF_{B,1})+2\tR_M(\fF^2_{B,1})\) and \begin{align}D\fF^2_{B,1}&:=\{\bar{\phi}: \bar{\phi}=(D_i\phi)^2,\phi\in\fF_{B,1},\text{ $D_i$ is the weak derivative in the $i$-th variable}\},\notag\\D\fF_{B,1}&:=\{\bar{\phi}: \bar{\phi}=D_i\phi,\phi\in\fF_{B,1},\text{ $D_i$ is the weak derivative in the $i$-th variable}\},\end{align}$\rmE$ is expected responding to $X$, and $X:=\{\vx_1,\ldots,\vx_M\}$ is an independent random variables set uniformly distributed on $[0,1]^d$.
			\end{proposition}

   \begin{proof}Due to the definition, we know that \[\fR_{D,1}(\vtheta_{D,1})=\inf_{\phi\in\fF_{B,1}}\|\phi-f\|_{H^{1}\left([0,1]^d\right)}.\] For the sample error, due to $\vtheta_{S,1}$ and $\vtheta_{D,1}$ belong to $\Theta$ almost surely, we have\begin{align}
					& \left|\rmE\left[\fR_{S,1}(\vtheta_{S,1})-\fR_{D,1}(\vtheta_{S,1})\right]\right|\notag\\\le &\left|\rmE\left[\frac{1}{M}\sum_{i=1}^M\left| \nabla f(\vx_i)-\nabla\phi(\vx_i;\vtheta_{S,1})\right|^2-\int_{[0,1]^d}\left|\nabla f(\vx)-\nabla\phi(\vx;\vtheta_{S,1})\right|^2\,\D \vx +\fE_0\right]\right|\notag\\\le&2\left|\rmE\left(\frac{1}{M}\sum_{i=1}^M\nabla f(\vx_i)\nabla\phi(\vx_i;\vtheta_{S,1})-\int_{[0,1]^d} \nabla f(\vx)\nabla\phi(\vx;\vtheta_{S,1})\,\D \vx\right)\right|\notag\\&+ \left|\rmE\left(\frac{1}{M}\sum_{i=1}^M\nabla\phi(\vx_i;\vtheta_{S,1})^2-\int_{[0,1]^d} \nabla\phi(\vx;\vtheta_{S,1})^2\,\D \vx\right)\right|+\fE_0\notag\\\le&2\tR_M(\nabla fD\fF_{B,1})+\tR_M(D\fF^2_{B,1})+\fE_0
				\end{align}where the last inequality is due to Lemma \ref{connect1} and $\fE_0=4\tR_M(\fF_{B,1})+2\tR_M(\fF^2_{B,1})$. 

    Due to Lemmas \ref{connect1} and \ref{jiao}, and $|\nabla f|\le d$, we have that \begin{align}
        &\rmE\left|\frac{1}{M}\sum_{i=1}^M\nabla f(\vx_i)\nabla\phi(\vx_i;\vtheta_{S,1})-\int_{[0,1]^d} \nabla f(\vx)\nabla\phi(\vx;\vtheta_{S,1})\,\D \vx)\right|\notag\\\le& 2d\tR_M(fD\fF_{B,1})\le 2d\tR_M(D\fF_{B,1}). 
    \end{align}
    Therefore, we have \begin{align}
        \rmE\fR_{S,1}(\vtheta_{S,1})-\fR_{D,1}(\vtheta_{S,1})\le \fE_1+\fE_0,
    \end{align}where $\fE_1=4d\tR_M(D\fF_{B,1})+2\tR_M(D\fF^2_{B,1})$.
			\end{proof}

   To further refine our understanding, we aim to estimate $\tR_M(D\fF_{B,1})$ and $\tR_M(D\fF^2_{B,1})$ in order to bound the generalization error. The pseudo-dimension can be used to provide such bounds. Notably, for DeNN derivatives, \cite{yang2023nearly} presents nearly optimal bounds for the pseudo-dimension. In the upcoming propositions, we will demonstrate the pseudo-dimension of $D\fF^2_{B,1}$.

    \begin{proposition}
        \label{pdimh1}
			For any $N,L,d\in\sN_+$, there exists a constant $\bar{C}$ independent with $N,L$  such that 	\begin{equation}
				\text{Pdim}(D\fF_{B,1}^2)\le \bar{C} N^2L^2\log_2 L\log_2 N.
			\end{equation}
		\end{proposition}
  \begin{proof}
  The proof is similar to those of Propositions \ref{pdimh11} and \ref{pdim}.
		\end{proof}

   \begin{proof}[Proof of Theorem \ref{generalzation l2R} for $k=1$]
       The proofs of Theorem \ref{generalzation l2R} for $k=1$ are analogous to those of Theorem \ref{generalzation l2R} for $k=0$ by combining the results of Propositions \ref{connect radh1R} and \ref{pdimh1}.
   \end{proof}

   \subsubsection{Proof of Theorem \ref{generalzation l2R} for $k=2$}


   \begin{proposition}
			\label{connect radh2R}
				Let $d, N, L,M\in\sN_+$, $B,C_1,C_2\in\sR_+$. For any $f\in W^{n,\infty}([0,1]^d)$ with $\|f\|_{W^{n,\infty}([0,1]^d)}\le 1$, we have \[\rmE\fR_{D,1}(\vtheta_{S,1})\le \inf_{\phi\in\fF_{B,2}}\|\phi-f\|_{H^{2}\left([0,1]^d\right)}+\fE_2+\fE_0,\]where \(\fE_2=4d\tR_M(D^2\fF_{B,2})+2\tR_M(D^2\fF^2_{B,2})\) and \(\fE_0=4\tR_M(\fF_{B,2})+2\tR_M(\fF^2_{B,2})\) and \begin{align}D^2\fF^2_{B,2}&:=\{\bar{\phi}: \bar{\phi}=(D^2_i\phi)^2,\phi\in\fF_{B,2},\text{ $D^2_i$ is the second weak derivative in the $i$-th variable}\},\notag\\D^2\fF_{B,2}&:=\{\bar{\phi}: \bar{\phi}=D^2_i\phi,\phi\in\fF_{B,2},\text{ $D^2_i$ is the second weak derivative in the $i$-th variable}\},\end{align}$\rmE$ is expected responding to $X$, and $X:=\{\vx_1,\ldots,\vx_M\}$ is an independent random variables set uniformly distributed on $[0,1]^d$.
			\end{proposition}

   \begin{proof}The proof is similar to those of Proposition \ref{connect radh1R}.
			\end{proof}

To further refine our understanding, we aim to estimate $\tR_M(D^2\fF_{B,2})$ and $\tR_M(D^2\fF^2_{B,2})$ in order to bound the generalization error. The pseudo-dimension can be used to provide such bounds. Notably, for DeNN derivatives, \cite{yang2023nearly} presents nearly optimal bounds for the pseudo-dimension. In the upcoming propositions, we will demonstrate the pseudo-dimension of $D^2\fF^2_{B,2}$.

    \begin{proposition}\label{pdimh2}
			For any $N,L,d\in\sN_+$, there exists a constant $\bar{C}$ independent with $N,L$  such that 	\begin{equation}
				\text{Pdim}(D^2\fF_{B,2}^2)\le \bar{C} N^2L^2\log_2 L\log_2 N.
			\end{equation}
		\end{proposition}	
  \begin{proof}
  The proof is similar to those of Propositions \ref{pdimh21} and \ref{pdim}.
		\end{proof}

  \begin{proof}[Proof of Theorem \ref{generalzation l2} for $k=2$]
       The proofs of Theorem \ref{generalzation l2} for $k=2$ are analogous to those of Theorem \ref{generalzation l2} for $k=0$ by combining the results of Propositions \ref{connect radh2R} and \ref{pdimh2}.
   \end{proof}
   \section{Experiment}\label{experiment}

\paragraph{Target function:} Here we define the target function as 
\begin{equation}
    f(\bm x): [0,1]^2 \mapsto \mathbb R
\end{equation}
with
\begin{equation}
    f(\bm x) = \left(\sum_{i=1}^{n} a_i\sigma(w_i \cdot x + b_i) \right) \times \left( \sum_{i=1}^{n} \alpha_i\sigma(\omega_i \cdot x + \beta_i) \right)
\end{equation}
where $\sigma = {\rm ReLU}$, $n=1000$, and $a_i, w_i, b_i, \alpha_i, \omega_i, \beta_i$ are randomly chosen by the default sampling strategies in PyTorch. As a result, 
\begin{equation}
  f(\bm x ) \in W^{1, \infty}\left([0,1]^2\right).  
\end{equation}

\paragraph{Large Data Regime:} In this scenario, both networks are trained and tested on datasets with $10,000$ points sampled from the domain $[0,1]\times[0,1]$. 

\paragraph{Small Data Regime:} In this case, the training and testing are performed on a smaller dataset of $1,000$ samples from the same domain.

\paragraph{Test and results:} We randomly sample 10,000 points for testing for both large and small data cases. Results shown in Table \ref{tab:my_label} are based on 3 tests. 

\begin{table}[h]
    \centering
    \begin{tabular}{|c|c|c|}
    \hline
       Neural Network  &  Large Data Regime & Small Data Regime \\
       \hline
       Shallow (Depth 1, Width 20)  & Mean: 6.18e-4, Std: 9.86e-05 & Mean: 1.51e-3, Std: 2.52e-4 \\
       \hline
       Deep (Depth 4, Width 10)      & Mean: 3.69e-4, Std: 2.57e-05 & Mean: 4.96e-3, Std: 4.46e-3 \\
       \hline
    \end{tabular}
    \caption{This table compares the performance of shallow and deep neural networks in terms of mean test performance and standard deviation across different data regimes. It illustrates how network depth and data availability impact learning outcomes, with shallow networks performing better in small data scenarios, while deep networks excel with larger datasets.}
    \label{tab:my_label}
\end{table}


\end{document}